\documentclass[letterpaper]{article} % DO NOT CHANGE THIS
\usepackage{aaai2026}  % DO NOT CHANGE THIS
\usepackage{times}  % DO NOT CHANGE THIS
\usepackage{helvet}  % DO NOT CHANGE THIS
\usepackage{courier}  % DO NOT CHANGE THIS
\usepackage[hyphens]{url}  % DO NOT CHANGE THIS
\usepackage{graphicx} % DO NOT CHANGE THIS
\usepackage{natbib}  % DO NOT CHANGE THIS AND DO NOT ADD ANY OPTIONS TO IT
\usepackage{caption} % DO NOT CHANGE THIS AND DO NOT ADD ANY OPTIONS TO IT
\usepackage{algorithm}
\usepackage{algorithmic}

\usepackage{newfloat}
\usepackage{listings}
\DeclareCaptionStyle{ruled}{labelfont=normalfont,labelsep=colon,strut=off} % DO NOT CHANGE THIS
\floatstyle{ruled}
\newfloat{listing}{tb}{lst}{}
\floatname{listing}{Listing}
\usepackage{graphicx}
\usepackage{algorithm}
\usepackage{algorithmic, mathrsfs, bm}
\usepackage{multirow}
\usepackage{array}
\usepackage{amsmath}
\usepackage{amssymb}
\usepackage{appendix}
\usepackage{footmisc}
\usepackage{colortbl}
\usepackage{cleveref}
\usepackage{booktabs}
\usepackage{amsthm}
\newtheorem{theorem}{Theorem}%  meant for continuous numbers

\newtheorem{lemma}{Lemma}
\newtheorem{assumption}{Assumption}
\usepackage{pifont}

\title{Efficiently Seeking Flat Minima for Better Generalization in Fine-Tuning Large Language Models and Beyond}

\author {
    Jiaxin Deng\textsuperscript{\rm 1}\equalcontrib,
    Qingcheng Zhu\textsuperscript{\rm 2}\equalcontrib,
    Junbiao Pang\textsuperscript{\rm 1}\thanks{Corresponding authors.},
    Linlin Yang\textsuperscript{\rm 3},
    Zhongqian Fu\textsuperscript{\rm 4},
    Baochang Zhang\textsuperscript{\rm 5}\thanks{Project leader.}
}
\affiliations {
    \textsuperscript{\rm 1}School of Information Science and Technology, Beijing University of Technology, Beijing, China\\
    \textsuperscript{\rm 2}School of Electronic Information Engineering
    , Beihang University, Beijing, China\\
    \textsuperscript{\rm 3}State Key Laboratory of Media Convergence and Communication, Communication University of China, Beijing, China\\
    \textsuperscript{\rm 4}Huawei Noah's Ark Lab, China\\
    \textsuperscript{\rm 5}Hangzhou Research Institute, School of Artificial Intelligence, Beihang University, China\\
    dengjiaxin@emails.bjut.edu.cn, zhuqc@buaa.edu.cn, junbiao\_pang@bjut.edu.cn, lyang@cuc.edu.cn, fuzhongqian@huawei.com, bczhang@buaa.edu.cn
}

\usepackage{bibentry}
\begin{document}

\maketitle

\begin{abstract}
Little research explores the correlation between the expressive ability and generalization ability of the low-rank adaptation (LoRA).
Sharpness-Aware Minimization (SAM) improves model generalization for both Convolutional Neural Networks (CNNs) and Transformers by encouraging convergence to locally flat minima. However, the connection between sharpness and generalization has not been fully explored for LoRA due to the lack of tools to either empirically seek flat minima or develop theoretical methods. 
In this work, we propose Flat Minima LoRA (FMLoRA) and its efficient version \textit{i.e.}, EFMLoRA, to seek flat minima for LoRA. Concretely, we theoretically demonstrate that perturbations in the full parameter space can be transferred to the low-rank subspace.
This approach eliminates the potential interference introduced by perturbations across multiple matrices in the low-rank subspace.  
Our extensive experiments on large language models and vision-language models demonstrate that EFMLoRA achieves optimize efficiency comparable to that of LoRA while simultaneously attaining comparable or even better performance. For example, on the GLUE dataset with RoBERTa-large, EFMLoRA outperforms
LoRA and full fine-tuning by 1.0\% and 0.5\% on average, respectively. On vision-language models \textit{e.g.}, Qwen-VL-Chat, there are performance improvements of 1.5\% and 1.0\% on the SQA and VizWiz datasets, respectively. These empirical results also verify that the generalization of LoRA is closely related to sharpness, which is omitted by previous methods. 
\end{abstract}


\section{Introduction}
Parameter-Efficient Fine-Tuning (PEFT) methods only update a small subset of parameters, \textit{e.g.}, adapters \cite{hu2022lora} or prompt weights \cite{li2021prefix} for Large language models (LLMs) with substantially lower memory and computational costs. Specifically, Low-Rank Adaptation (LoRA) \cite{hu2022lora} stands out for achieving performance comparable to full fine-tuning (FT) while being considerably more efficient.
\begin{figure}[t]
\centering
\includegraphics[width=0.8\columnwidth]{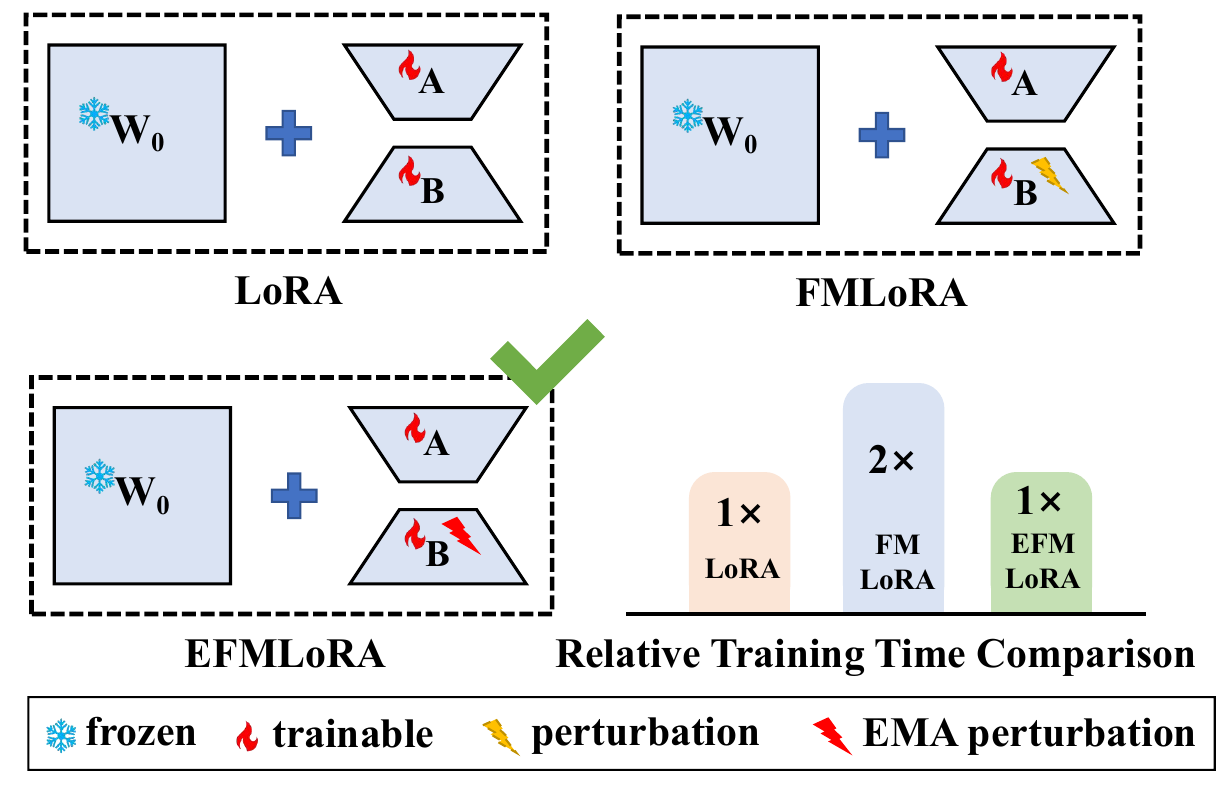} 
\caption{Comparison of Methods: LoRA, FMLoRA, and EFMLoRA.}
\label{fig_comp}
\end{figure}

Many works have been proposed to enhance the performance of LoRA by introducing more dedicated budgets for rank
allocation \cite{zhang2023adaptive}, decomposing optimization for direction and magnitude updates \cite{liu2024dora}, or designing better initialization strategies for LoRA parameters \cite{meng2024pissa}, etc. These studies demonstrate the significant potential to improve LoRA performance. However, most existing approaches fail to effectively address bias inheritance, where LLMs may propagate and amplify their inherent biases, significantly impacting model performance and robustness on downstream tasks \cite{li2025understanding}. Therefore, a natural question is: how to model and understand the generalization of LoRA for various LLMs and beyond, \textit{e.g.}, vision-language models?

%SAM介绍
It is widely believed that a flatter loss landscape can lead to better generalization performance~\cite{hochreiter1994simplifying}~\cite{hochreiter1997flat}. For instance, Foret et al. proposed Sharpness-Aware Minimization (SAM)~\cite{foret-2020-SAM-ICLR}, which seeks parameter regions where the training loss remains uniformly flat. SAM and its variants have demonstrated State-Of-The-Art (SOTA) performances across various applications, such as classification~\cite{kwon-2021-asam-ICML}, transfer learning~\cite{zhuang-2022-GSAM-ICLR}, domain generalization~\cite{dong2024implicit} and federated learning~\cite{FedGAMMA}.

To the best of our knowledge, compared to theoretical analysis, \textit{e.g.},~\cite{neyshabur2017exploring}, empirically connecting sharpness and generalization ability of LoRA is a practical approach, \textit{e.g.},~\cite{andriushchenko2023modern}. For the second line of research, a naive approach is to combine SAM with LoRA.
However, if perturbations in SAM are applied simultaneously to two low-rank subspaces of LoRA, they may change the maximum loss within the neighborhood of LoRA’s full parameter space~\cite{dinh2017sharp}; besides, SAM incurs a computational cost twice that of Stochastic Gradient Descent (SGD)~\cite{deng2024effective}. The key question in the second line of research is how to efficiently find flat minima in LoRA, aiming to better understand the connection between sharpness and generalization.

In this paper, we propose a novel PEFT method, FMLoRA, that promotes convergence toward flatter minima. Specifically, we theoretically uncover that  perturbations in the full parameter space can be equivalently re-parameterized as perturbations within the low-rank space. In addition, we propose EFMLoRA to accelerate FMLoRA by an Exponential Moving Average (EMA) strategy. 
We validate that EFMLoRA improves generalization performance on downstream tasks while maintaining computational efficiency comparable to that of LoRA. Fig.\ref{fig_comp} compares three methods: LoRA, FMLoRA, and EFMLoRA. We conducted comprehensive experiments on diverse tasks (fine-tuning, few-shot learning) and various model types (RoBERTa \cite{liu2019roberta}, GPT-2 \cite{radford2019language}, CLIP \cite{zanella2024low}, Qwen-VL-Chat \cite{Bai2023QwenVLAV}) and scales. 
We find that EFMLoRA achieves model accuracy very close to, or even surpass both full fine-tuning and LoRA across many tasks. Our main contribution can be summarized as follows:
\begin{itemize}
    \item We propose FMLoRA, a novel PEFT training method that integrates SAM into the LoRA framework. Furthermore, EFMLoRA provides an efficient tool for empirically understanding the connection between sharpness and generalization in LLMs and beyond. We empirically show that reducing sharpness is highly correlated with improved generalization in PEFT tasks, which has been rarely explored in PEFT studies before.
    \item We conduct comprehensive experiments on LLMs (\textit{e.g.}, RoBERTa, GPT-2) and vision-language models (\textit{e.g.}, CLIP, Qwen-VL-Chat) across various tasks including fine-tuning and few-shot learning. Results show that EFMLoRA achieves optimize efficiency comparable to that of LoRA while simultaneously attaining comparable or even better performance.
\end{itemize}

\section{Related Works}
\subsection{Low-rank Adaption}
Hu et al. proposed LoRA \cite{hu2022lora} as a PEFT method that introduced low-rank adapters into each layer of a pre-trained model. 
Recent advancements in LoRA can be broadly categorized into two directions: 1) advanced architectures and 2) optimization methods. In the first research line, for example, LoraHub \cite{huang2023lorahub} trained multiple adapters and strategically combined them based on the domain during inference. 
LoRA-FA \cite{zhang2023lora} chose to freeze the projection-down weight of $\mathbf{A}$ and update the projection-up weight of $\mathbf{B}$ in each LoRA layer. 
DoRA \cite{liu2024dora} improved LoRA by incorporating a learnable magnitude vector to re-scale the normalized product of low-rank matrices. 
HydraLoRA \cite{tian2024hydralora} extended the LoRA framework with an asymmetric architecture that shared a common $\mathbf{A}$ matrix for efficiency while dynamically assigning samples to multiple $\mathbf{B}$ matrices via a MoE mechanism.
In the second line, for example, 
LoRA+ \cite{hayou2024lora+} applied different learning rates to the two low-rank matrices. 
Additionally, Galore \cite{zhao2024galore} employed SVD to compress the gradients and its first and second momentum of full training into a low-rank space, thereby reducing the memory footprint during pre-training and fine-tuning. 
Recently, Li et al. \cite{li2024flat} proposed combining SAM with LoRA for better generalization, but they used random perturbation.
Our method belongs to the second research line. Different from \cite{li2024flat}, our method transfers the perturbation from the full parameter space to a single low-rank parameter space without changing
the maximum perturbed loss, avoiding misalignment with SAM's training behavior.

\subsection{Sharpness and Generalization Ability}
Research on the relationship between sharpness and generalization could be traced back to \cite{hochreiter1997flat}. Following the observation by \cite{keskar-2016-large_batch-ICLR} that larger batch sizes tended to increase sharpness and generalization error. \cite{jastrzkebski2017three} extended this by finding a correlation between the sharpness and the ratio of learning rate to batch size. \cite{dinh-2017-sharp_minima-ICML} showed that one can easily construct networks with good generalization but with arbitrary large sharpness
by reparameterization. \cite{jiang-2019-fantastic-ICLR} performed a large-scale empirical study on various generalization measures and showed that sharpness-based measures have the highest correlation with generalization. Theoretical understandings on the generalization error using sharpness-related measures were provided in \cite{neyshabur2017exploring}, \cite{wanggeneralization}. Collectively, these studies justified the goal of seeking flatter minima to improve generalization. However, to the best of our knowledge, the correlation between sharpness and generalization for LoRA has barely been discussed due to the lack of theoretical understanding or efficient tools for empirical analysis. Our method provides an efficient tool for empirical analysis in this domain.

\subsection{Recap of SAM}
Foret et al.~\cite{foret-2020-SAM-ICLR} proposed the SAM to enhance model generalization as follows:
\begin{equation}\label{equ:sam_max}
\begin{aligned}
\mathop {\min } \limits_\mathbf{w} [(\mathop {\max }\limits_{||\bm{\varepsilon} || \le \rho } L(\mathbf{w} + \bm{\varepsilon} )- L(\mathbf{w}) ) + L(\mathbf{w}) + \lambda ||\mathbf{w}||_2^2],
\end{aligned}
\end{equation}
where $\mathbf{w}$ represents the weights of the network, $\bm{\varepsilon}$ represents the perturbation of weights $\mathbf{w}$ in a Euclidean ball with the radius $\rho$ $(\rho>0)$, $L(\cdot)$ is the loss function, and $\lambda ||\mathbf{w}||_2^2$ is a standard L2 regularization term.

SAM utilizes Taylor expansion to search for the maximum perturbed loss ($\mathop {\max }\limits_{||\bm{\varepsilon} || \le \rho } L(\mathbf{w} + \bm{\varepsilon} )$) in local parameter space as follows:
\begin{equation}\label{equ:psf}
\begin{aligned}
\mathop {\arg \max }\limits_{||\bm{\varepsilon} || \le \rho } \; L(\mathbf{w} + \bm{\varepsilon} ) 
\approx \mathop {\arg \max }\limits_{||\bm{\varepsilon} || \le \rho } \; {\bm{\varepsilon} ^{\top}}{\nabla _\mathbf{w}}L(\mathbf{w}).
\end{aligned}
\end{equation}
By solving Eq.~\eqref{equ:psf}, SAM obtains the perturbation as follows:
\begin{equation}\label{equ:e_max}
\begin{aligned}
\hat{ \bm{\varepsilon} } = \rho {\nabla _\mathbf{w}}L(\mathbf{w})/||{\nabla _\mathbf{w}}L(\mathbf{w})||.
\end{aligned}
\end{equation}
Substituting the perturbation $\hat{\bm{\varepsilon} }$ back into Eq.~\eqref{equ:sam_max}, we then have:
\begin{equation}\label{equ:gs}
\begin{aligned}
&{\nabla _{\mathbf{w}}}\mathop {\max }\limits_{||\bm{\varepsilon} || \le \rho } L(\mathbf{w} + \bm{\varepsilon} ) \approx {\nabla _{\mathbf{w}}}L({\mathbf{w}} + \hat {\bm{\varepsilon}} ({\mathbf{w}})) \\ 
&= {\nabla _{\mathbf{w}}}L({\mathbf{w}}){|_{{\mathbf{w}} + \hat {\bm{\varepsilon}} ({\mathbf{w}})}} + \frac{{d\hat {\bm{\varepsilon}} ({\mathbf{w}})}}{{d{\mathbf{w}}}}{\nabla _{\mathbf{w}}}L({\mathbf{w}}){|_{{\mathbf{w}} + \hat {\bm{\varepsilon}} ({\mathbf{w}})}}.
\end{aligned}
\end{equation}
By dropping the second-order terms in Eq.\eqref{equ:gs}, SAM calculates the gradient at $\mathbf{w}+\bm{\hat\varepsilon}$ as follows:
\begin{equation}\label{equ:grad_sam}
\begin{aligned}
{\nabla _{\mathbf{w}}}\mathop {\max }\limits_{||\bm{\varepsilon} || \le \rho } L(\mathbf{w} + \bm{\varepsilon} ) \approx {\nabla _\mathbf{w}}L(\mathbf{w}){|_{\mathbf{w} + \bm{\hat\varepsilon} }}.
\end{aligned}
\end{equation}
Finally, SAM uses the gradients from Eq.~\eqref{equ:grad_sam} for optimization.

\subsection{SAM Variants}
Recently, SAM variants could be broadly categorized into three groups: 1) studies on the perturbation radius $\rho$ in SAM, 2) studies that speed up the optimization process of SAM, and 3) redefinitions of sharpness in SAM. For the first direction, Kwon et al.~\cite{kwon-2021-asam-ICML} proposed Adaptive SAM (ASAM), which adapted the perturbation radius in a scale-aware manner, allowing SAM to be effectively applied to scale-invariant neural networks.
For the second group, Kim et al.~\cite{kim2023exploring} introduced a multi-step ascent approach to improve SAM. Li et al.~\cite{li2024friendly} introduced Friendly SAM (F-SAM), which improved generalization by removing the detrimental influence of the full gradient component and instead utilizing batch-specific gradients to guide optimization more effectively.
For the third group, Zhuang et al.~\cite{zhuang-2022-GSAM-ICLR} pointed out that SAM did not always favor flat minima. Consequently, they proposed GSAM, which minimized the surrogate gap and the perturbed loss to better encourage flatness. Zhang et al. introduced the first-order flatness~\cite{zhang-2023-gradient-CVPR}, which assessed the maximal gradient norm within a perturbation radius. Consequently, they proposed GAM which explicitly seeks minima characterized by uniformly small curvature.

\section{Method}
\subsection{SAM on LoRA}
LoRA achieves parameter efficiency by modeling the low-rank decomposed weight~\cite{li2022low}. Specifically, the weight change for each layer $\mathbf{W}_0 \in \mathbb{R}^{n \times m}$ is represented as $\Delta \mathbf{W} = s \mathbf{B} \mathbf{A}$, where $s$ is a scaling factor, $\mathbf{B} \in \mathbb{R}^{n \times r}$,  $\mathbf{A} \in \mathbb{R}^{r \times m}$, with rank $r \ll \min(n, m)$. Given an input $\mathbf{x}$, the forward is as follows:
\begin{align}
\mathbf{y}=\mathbf{W}_0\mathbf{x}+\Delta \mathbf{W}\mathbf{x}=(\mathbf{W}_0+s \mathbf{B}\mathbf{A})\mathbf{x},
\end{align}
where matrix $\mathbf{A}$ is typically initialized by the Kaiming's method~\cite{he2015delving}, $\mathbf{B}$ is set to zeros. $\mathbf{W}_0$ remains unchanged during fine-tuning, while $\mathbf{B}$ and $\mathbf{A}$ are trained. During inference, $\Delta \mathbf{W}$ is merged into $\mathbf{W_0}$.

If SAM is naively combined with LoRA, the optimization loss can be rewritten as follows:
\begin{align}\label{equ:dir_pert}
\min_{\mathbf{A},\mathbf{B}} ~~\mathop {\max }\limits_{\scriptstyle||{{\bf{E}}^{\bf{A}}}|{|_F} \le \rho ,\hfill\atop
\scriptstyle||{{\bf{E}}^{\bf{B}}}|{|_F} \le \rho \hfill}L({\mathbf{W_0}} + {s}(\mathbf{B} +  {\mathbf{E}^{\mathbf{B}}})(\mathbf{A} +  {\mathbf{E}^\mathbf{A}})),
\end{align}
where $\mathbf{E}^{\mathbf{B}}\in \mathbb{R}^{n \times r}$ and $\mathbf{E}^{\mathbf{A}}\in \mathbb{R}^{r \times m}$ represent the perturbations applied to the parameters $\mathbf{B}$ and $\mathbf{A}$, respectively, and $\rho$ is the radius of perturbations. There are two key challenges: 
\begin{itemize}
    \item Two separate perturbations in two low-rank subspaces interfere with each other, leading to an inconsistency between the maximum loss obtained when perturbing in the low-rank subspaces and the maximum loss obtained when perturbing in the full parameter space.
    \item SAM requires computing gradients twice per iteration, resulting in approximately twice the computational cost compared to LoRA.
\end{itemize}

\subsection{FMLoRA}
To deal with the first challenge, we propose to re-parameterize the perturbation from the full parameter space to a single low-rank parameter space. Concretely, the loss in the full parameter space can be formulated as follows:
\begin{align}\label{equ:full_spa}
\min_{\mathbf{A},\mathbf{B}} ~~\max_{\|\mathbf{E}^\mathbf{W}\|_F\le \rho}~~ L(\mathbf{W_0}+s \mathbf{B}\mathbf{A}+\mathbf{E}^\mathbf{W}).
\end{align}
To solve the minimax problem in Eq.~\eqref{equ:full_spa}, it is necessary to first find optimal $\hat{\mathbf{E}}^\mathbf{W} \in \mathbb{R}^{n \times m}$. Analogous to SAM, we approximate the optimal perturbation $\hat{\mathbf{E}}^\mathbf{W}$ to maximize $L(\mathbf{W}+\mathbf{E}^\mathbf{W})$ where $\mathbf{W}=\mathbf{W_0}+s \mathbf{B}\mathbf{A}$ as follows:
\begin{align}\label{pertur}
\hat{\bm{\varepsilon}}^\mathbf{w}=\rho \text{sign}(\mathbf{g}^{\mathbf{w}}) \frac{\mathbf{g}^{\mathbf{w}}}{||\mathbf{g}^{\mathbf{w}}||},  
\end{align}
where $\mathbf{g}^{\mathbf{w}}=\text{Vector}(\nabla L_\mathbf{W}(\mathbf{W}))$ and $\hat{\bm{\varepsilon}}^\mathbf{w}=\text{Vector}(\hat{\mathbf{E}}^\mathbf{W})$, in which the $\text{Vector}(\cdot)$ function represents a vectorized operation. However, the solution for $\hat{\mathbf{E}}^\mathbf{W}$ explicitly depends on the gradient of the matrix $\mathbf{W}$. That is, the form of solution in~Eq.~\eqref{pertur} is undesirable since $\nabla L_\mathbf{W}(\mathbf{W})$ is unknown during LoRA optimization. 

In this paper, we propose to approximate the unknown gradient $\nabla L_\mathbf{W}(\mathbf{W})$ using standard LoRA gradients, which can be computed in two ways:
\begin{align}\label{eq:two_appro}
(1)\quad \nabla L_\mathbf{W}(\mathbf{W})= \frac{1}{s}  \nabla L_\mathbf{B}(\mathbf{W_0}+ s\mathbf{BA})(\mathbf{A}^\top)^+, \\
(2)\quad \nabla L_\mathbf{W}(\mathbf{W})=\frac{1}{s} (\mathbf{B}^\top)^+\nabla L_\mathbf{A}(\mathbf{W_0}+ s\mathbf{BA}),
\end{align}
where $(\mathbf{A}^\top)^+$ and $(\mathbf{B}^\top)^+$ represent the pseudo-inverse of $\mathbf{A}^\top$ and $\mathbf{B}^\top$, respectively. The accuracy of the pseudo-inverse depends on the condition number of matrix. A smaller condition number leads to a more accurate pseudo-inverse. Matrices with lower condition numbers are better suited for stable representation. In LoRA, we found that the condition number is typically low, around 3. 

To obtain a more accurate estimate of the gradient of the full weights, we combine the above two approaches to compute $\nabla L_\mathbf{W}(\mathbf{W})$ as follows:
\begin{align}
\overline{\nabla {L}_\mathbf{W}(\mathbf{W})}&=0.5*(\frac{1}{s}  \nabla L_\mathbf{B}(\mathbf{W_0}+ s\mathbf{BA})(\mathbf{A}^\top)^+ \nonumber \\ &+ \frac{1}{s} (\mathbf{B}^\top)^+\nabla L_\mathbf{A}(\mathbf{W_0}+ s\mathbf{BA})).
\end{align}
Let ${\Bar{\mathbf{g}}^\mathbf{W}}=\text{Vector}(\overline{\nabla {L}_\mathbf{W}(\mathbf{W})})$.  Then the perturbation in Eq.~\eqref{pertur} could be rewritten as follows:
\begin{align}
%\Bar{\bm{\varepsilon}}^\mathbf{w}=\text{Vector}(\Bar{\mathbf{E}}^\mathbf{W})=
\Bar{\mathbf{E}}^\mathbf{W}=\text{Matrix}(
\rho \text{sign}({\Bar{\mathbf{g}}^\mathbf{W}}) \frac{{\Bar{\mathbf{g}}^\mathbf{W}}}{||{\Bar{\mathbf{g}}^\mathbf{W}}||}),  
\end{align}
where $\text{Matrix}(\cdot)$ denotes the operation that converts a vector into a matrix.
We transfer the perturbation from the full parameter space to a single low-rank parameter space without changing the maximum loss in the local region of the parameters. We apply no perturbation to matrix $\mathbf{A}$, \textit{i.e.}, ${\mathbf{E}^\mathbf{A}} = \mathbf{0}$, and ensure that the loss under perturbations in the low-rank subspace in Eq.~\eqref{equ:dir_pert} matches the inner maximum loss in Eq.~\eqref{equ:full_spa}, as follows:
\begin{equation}\label{equ:loss_match}
\begin{aligned}
L({\mathbf{W_0}} &+ {s}(\mathbf{B} + 
{\mathbf{E}^{\mathbf{B}}})\mathbf{A})\\&=\max_{\|\mathbf{E}^\mathbf{W}\|_F\le \rho} L(\mathbf{W_0}+s \mathbf{B}\mathbf{A}+\mathbf{E}^\mathbf{W}).
% \min_{\mathbf{A},\mathbf{B}} ~~\mathop {\max }\limits_{\scriptstyle
% \scriptstyle||{{\bf{E}}^{\bf{B}}}|{|_F} \le \rho \hfill}L({\mathbf{W_0}} + {s}(\mathbf{B} +  {\mathbf{E}}^{\mathbf{B}}})\mathbf{A}),\\
% {\mathbf{W_0}} + {s}(\mathbf{B} +  {\Bar{\mathbf{E}}^{\mathbf{B}}})\mathbf{A}\approx\mathbf{W_0}+s \mathbf{B}\mathbf{A}+\Bar{\mathbf{E}}^\mathbf{W}
\end{aligned}
\end{equation}
Substituting $\Bar{\mathbf{E}}^\mathbf{W}$ into Eq.~\eqref{equ:loss_match}, we obtain:
\begin{align}\label{equ:EB}
{\mathbf{E}}^\mathbf{B}\approx\frac{1}{s}\Bar{\mathbf{E}}^\mathbf{W}\mathbf{A}^+,
\end{align}
% $s{\mathbf{E} ^{\mathbf{B}}}\mathbf{A}=\Bar{\mathbf{E}}^\mathbf{W}$ and $s\mathbf{B}{\mathbf{E} ^\mathbf{A}}=\Bar{\mathbf{E}}^\mathbf{W}$, where ${\mathbf{E} ^{\mathbf{B}}}$ and ${\mathbf{E} ^{\mathbf{A}}}$ represent perturbations applied to matrices $\mathbf{B}$ and $\mathbf{A}$, respectively.
% Based on Eq.~\eqref{equ:full_spa}, the perturbation $\Bar{\mathbf{E}}^\mathbf{W}$ can be transferred to either matrix $\mathbf{B}$ or $\mathbf{A}$, resulting in the following two cases:
% \begin{align}
% (1) \quad \min_{\mathbf{A},\mathbf{B}}  L(\mathbf{W_0}+s\cdot (\mathbf{B}+\frac{1}{s}\Bar{\mathbf{E}}^\mathbf{W}\mathbf{A}^+)\mathbf{A}) \label{equ:trans_B}, \\
% (2) \quad \min_{\mathbf{A},\mathbf{B}} L(\mathbf{W_0}+s\cdot \mathbf{B}(\mathbf{A}+\frac{1}{s}\mathbf{B}^+\Bar{\mathbf{E}}^\mathbf{W})),
% \end{align}
where $\mathbf{A}^+$ is the pseudo-inverse of $\mathbf{A}$. An alternative approach is to transfer the perturbation to matrix $\mathbf{A}$.
Following the observations from HydraLoRA~\cite{tian2024hydralora}, matrix $\mathbf{A}$ shows high parameter similarity across heads, likely due to initialization, making it capture domain-common features, while matrix $\mathbf{B}$ remains distinct and domain-specific. Since different tasks require different perturbations, we adopt the approach of transferring the perturbation to the matrix $\mathbf{B}$, as expressed in Eq.~\eqref{equ:loss_match}. The detailed derivation of Eq.~\eqref{eq:two_appro} and the pseudo-algorithm for FMLoRA are provided in the supplementary file.

\subsubsection{Balancedness of FMLoRA.}
Balancedness is well-appreciated in domains such as matrix factorization/sensing \cite{ge2017no} \cite{du2018algorithmic}.
It is also observed that balanced neural networks are easier to optimize relative to unbalanced ones \cite{neyshabur2015path}.
Recently, Balancedness $B_t:=\frac{1}{2}(||\mathbf{x}_t||^2-||\mathbf{y}_t||^2)$ (where $\mathbf{x}_t$ and $\mathbf{y}_t$ are variables) turns out to be an intriguing alternative to sharpness on the scale-invariant problem \cite{li2024implicit}.

To investigate the balancedness of our proposed method, we express the update process of FMLoRA analogously to Eq.(4) in \cite{li2024implicit} as follows:
\begin{equation}\label{equ:}
\begin{aligned}
{\tilde{\mathbf{x}}_t} = {{\mathbf{x}}_t} + \rho \frac{1}{s}\frac{{{\mathbf{G}_t}}}{{\left\| {{\mathbf{G}_t}} \right\|}}{\mathbf{y}_t}^+&, \quad {\tilde{\mathbf{y}}_t} = {{\mathbf{y}}_t},\\
{\mathbf{g}_{{\tilde{\mathbf{x}}_t}}} = {{\tilde{\mathbf{G}}}_t}\tilde{\mathbf{y}}_t&, \quad{\mathbf{g}_{{\tilde{\mathbf{y}}_t}}} = {{\tilde{\mathbf{G}}}_t}^{\top}\tilde{\mathbf{x}}_t,\\
{{\mathbf{x}}_{t + 1}} = {{\mathbf{x}}_t} - \eta {\mathbf{g}_{{\tilde{\mathbf{x}}_t}}}&, \quad{{\mathbf{y}}_{t + 1}} = {{\mathbf{y}}_t} - \eta {\mathbf{g}_{{\tilde{\mathbf{y}}_t}}},
\end{aligned}
\end{equation} 
where ${\mathbf{x}}_t=\text{Vector}(\mathbf{B}_t)$, ${\mathbf{y}}_t=\text{Vector}(\mathbf{A}_t)$, ${\mathbf{G}_t}=\nabla L({\mathbf{x}}_t{\mathbf{y}}_t^{\top})$ is the gradient of the full parameter space at the original parameter point, ${\tilde{\mathbf{G}}_t}=\nabla L(\tilde{\mathbf{x}}_t\tilde{\mathbf{y}}_t^{\top})$ is the gradient of the full parameter space at the perturbed parameter point, and $\mathbf{y}_t^+$ is the pseudo inverse of $\mathbf{y}_t$.

\begin{theorem}\label{thm:balan}
Let $B_t:=\frac{1}{2}(||\mathbf{x}_t||^2-||\mathbf{y}_t||^2)$. For the learning rate $\eta\Rightarrow 0$, the limiting flow of FMLoRA guarantees that:
\begin{equation}\label{equ:}
\begin{aligned}
\left| {\frac{1}{2}\frac{{d({{\left\| {{\mathbf{x}_t}} \right\|}^2} - {{\left\| {{\mathbf{y}_t}} \right\|}^2})}}{{dt}}} \right| \le \left| {\rho \frac{1}{s}\frac{1}{{\left\| {\mathbf{y}_t} \right\|}}\left\| {{\mathbf{g}_{{{{\rm{\tilde{\mathbf{x}}}}}_t}}}} \right\|} \right|.
\end{aligned}
\end{equation} 

\end{theorem}
Theorem 1 indicates that the balancedness of FMLoRA is influenced by the perturbation range $\rho$, the norm of the gradient at the perturbed point, the $\ell_2$-norm of $\mathbf{y}_t$, and the scale constraint of LoRA. To ensure that the balancedness of FMLoRA gradually decreases during training, we reduce $\rho$ progressively. In addition, the norm of the gradient with respect to $\mathbf{y}_t$ at the perturbed point also decreases due to the weight decay. The $\ell_2$-norm of $\mathbf{y}_t$ is bounded within a certain range, these factors collectively contribute to the reduction in the balancedness of FMLoRA.

\begin{figure}[t]
\centering
\includegraphics[width=0.8\columnwidth]{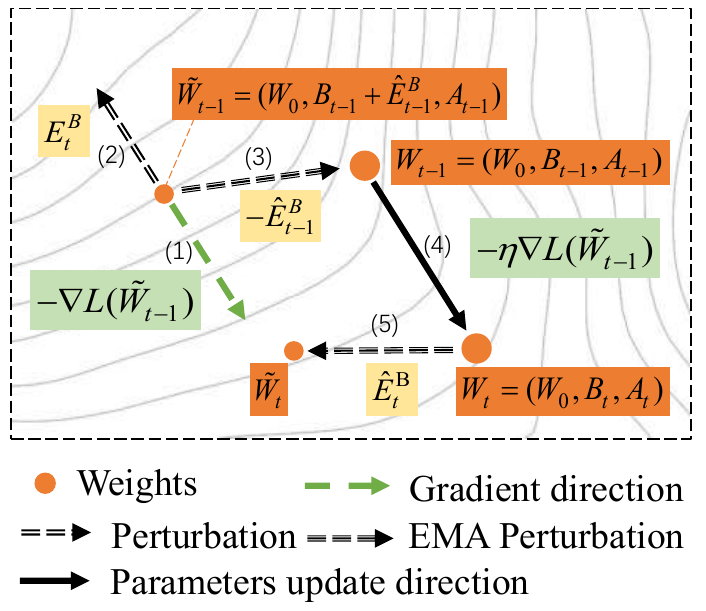} 
\caption{Parameter update process for EFMLoRA.}
\label{fig1}
\end{figure}

\subsection{Efficient FMLoRA}
The optimization processes of FMLoRA also require two gradient computations per iteration. 
To enhance optimization efficiency, we propose Efficient FMLoRA (EFMLoRA), which estimates the subsequent perturbation ${\mathbf{E}}^\mathbf{B}$ in Eq.~\eqref{equ:EB} by maintaining an Exponential Moving Average (EMA) of previous perturbations as follows:
\begin{equation}\label{equ:ema}
\begin{aligned}
{{\hat{\mathbf{E}} }^\mathbf{B}_t} = (1 - \beta ){{\hat{\mathbf{E}}}^\mathbf{B}_{t - 1}} + \beta {\mathbf{E}^\mathbf{B}_t},
\end{aligned}
\end{equation} 
where $\beta \in (0, 1)$ is the momentum coefficient that determines the update rate of the exponential moving average. ${\mathbf{E}^\mathbf{B}_t}$ is the perturbation on matrix $\mathbf{B}_t$ at $t$-th iteration, ${{\hat{\mathbf{E}} }^\mathbf{B}_t}$ is the EMA perturbation at $t$-th iteration. Fig.~\ref{fig1} illustrates the parameter update process of EFMLoRA: (1) Calculate the gradient at the perturbed  point $(\mathbf{W}_0$,  $\mathbf{B}_{t-1}+\hat{\mathbf{E}}^{\mathbf{B}}_{t-1}$, $\mathbf{A}_{t-1})$. (2) Calculate the perturbation ${\mathbf{E}}^\mathbf{B}_{t}=\frac{1}{s}\Bar{\mathbf{E}}^\mathbf{W}\mathbf{A}^+_{t-1}$. (3) Return to the original parameter point $(\mathbf{W}_0, \mathbf{B}_{t-1}, \mathbf{A}_{t-1})$. (4) Update the parameters to $(\mathbf{W}_0,  \mathbf{B}_t, \mathbf{A}_t)$. (5) Calculate the EMA perturbation by Eq.\eqref{equ:ema} and update the parameters to the next perturbed point $(\mathbf{W}_0, \mathbf{B}_t+\hat{\mathbf{E}}^{\mathbf{B}}_t, \mathbf{A}_t)$. During this optimization process, each optimization step requires only a single forward and backward. The algorithmic pseudocode is provided in the supplementary file.

\begin{table*}[]
\centering
\begin{tabular}{l|ccccccc}
\hline
RoBERTa    & SST-2   & SST-5   & SNLI    & MNLI    & RTE     & TREC  & avg.$\uparrow$  \\ \hline
Zero-Shot$*$  & 79.0    & 35.5    & 50.2    & 48.8    & 51.4    & 32.0 & 49.5    \\ \hline
LoRA$^*$       & 91.1$_{\pm0.8}$ & 52.3$_{\pm2.9}$ & 84.3$_{\pm0.3}$ & 78.1$_{\pm1.3}$ & 77.5$_{\pm2.3}$ & 96.6$_{\pm1.0}$ & 80.0 \\
LoRA-SAM$^*$   & \underline{92.2$_{\pm0.4}$} & 54.2$_{\pm2.0}$ & {85.5$_{\pm0.7}$} & {78.7$_{\pm1.0}$} & 80.6$_{\pm4.3}$ & \underline{96.7$_{\pm0.2}$} & 81.3 \\
LoRA-oBAR$^*$  & 91.5$_{\pm0.9}$ & 54.5$_{\pm2.7}$ & 84.9$_{\pm0.5}$ & 78.3$_{\pm2.2}$ & 79.7$_{\pm2.0}$ & \underline{96.7$_{\pm0.5}$} & 80.9 \\
LoRA-nBAR$^*$  & 91.4$_{\pm0.5}$ & \bf{55.0$_{\pm2.0}$} & 84.9$_{\pm1.4}$ & 78.1$_{\pm0.2}$ & {81.0$_{\pm1.0}$} & \underline{96.7$_{\pm1.0}$} & 81.2 \\
FMLoRA & \bf{95.1$_{\pm0.5}$} & {54.4$_{\pm1.3}$} & \bf{86.4$_{\pm0.8}$} & \bf{82.7$_{\pm1.0}$} & \underline{82.7$_{\pm1.2}$} & \underline{96.7$_{\pm0.2}$} & \bf{83.1} \\
EFMLoRA & {91.9$_{\pm1.7}$} & \underline{54.7$_{\pm1.6}$} & \underline{85.7$_{\pm0.7}$} & \underline{82.1$_{\pm0.6}$} & \bf{82.8$_{\pm0.2}$} & \bf{96.8$_{\pm0.4}$} & \underline{82.3} \\ \hline
\end{tabular}
\caption{Experiments on few-shot RoBERTa (355M). Results marked with $*$ are taken from \cite{li2024implicit}.} 
\label{fs_roberta} 
\end{table*}

To theoretically analyze the error of EFMLoRA, some necessary assumptions are listed below, all of which are common and standard when analyzing SAM optimization~\cite{du-2022-ESAM-ICLR}~\cite{zhuang-2022-GSAM-ICLR}.
\begin{assumption}\label{assumpion:smooth}
(Smooth) $L(\mathbf{w})$ is $\tau$-Lipschitz smooth in $\mathbf{w}$, \textit{i.e.}, $\left\| {\nabla L(\mathbf{w}) - \nabla L(\mathbf{v})} \right\| \le \tau \left\| {\mathbf{w} - \mathbf{v}} \right\|$.
\end{assumption}
\begin{assumption}\label{assumpion:Bounded_g}
(Bounded gradients). By the assumption that an upper bound exists on the gradient of each mini-batch. There exists $G > 0$ for each mini-batch such that $\mathbb{E}\left[ {\left\| {\nabla L(\mathbf{w})} \right\|} \right] \le G$.
\end{assumption}
\begin{assumption} \label{assumpion:bound_var}
(Bounded variance of stochastic gradients). Given the training set $\mathbf{D}$ and a mini-batch $\mathbf{B} \in \mathbf{D}$. There exists $\sigma \ge 0$, the variance of stochastic gradient $L_{\mathbf{B}}(\mathbf{w})$ is bounded by $\mathbb{E}\left[ {{{\left\| {\nabla {L_{\mathbf{B}}}(\mathbf{w}) - \nabla {L_\mathbf{D}}(\mathbf{w})} \right\|}^2}} \right] \le \sigma ^2$.
\end{assumption}
\begin{assumption} \label{assumpion:convex}
(Convex) We assume that the loss function $f : \mathbb{R}^n \rightarrow \mathbb{R}$ is convex and twice differentiable over an open domain. That is, for all $x, y \in \text{dom}(f)$, it satisfies:
$f(y) \geq f(x) + \nabla f(x)^\top (y - x)$.
\end{assumption}
This convexity assumption is reasonable in the fine-tuning stage, as the model is typically close to a local minimum and the loss landscape is approximately convex in a local neighborhood \cite{jang2024lora}.

\begin{theorem}\label{thm:sharpness}[EMA perturbation approximate perturbation of SAM due to the convex of the loss landscape]
Assume that during fine-tuning, the solution is already close to a local minimum and the local loss function is convex. Let the model weights at $i$-th iteration be $\mathbf{w}_t$. Under Assumptions \ref{assumpion:smooth}, \ref{assumpion:Bounded_g}, and \ref{assumpion:bound_var}, let ${\rho _t} = \frac{{{\rho _0}}}{{\sqrt t }}$, the error between the sharpness calculated using the EMA perturbation ($S^{\text{EMA}}$) and that calculated using the original SAM perturbation ($S^{\text{SAM}}$) is bounded as follows:
\begin{equation}\label{equ:}
\begin{aligned}
&|\underbrace {\left[ {L({\mathbf{w}_t} + {\bm{\hat \varepsilon }_{t}}) - L({\mathbf{w}_t})} \right]}_{S^{\text{EMA}}} - \underbrace {\left[ {L({\mathbf{w}_t} + {\bm{\tilde \varepsilon }_t}) - L({\mathbf{w}_t})} \right]}_{S^{\text{SAM}}}| \\
&\le \left( {\left( {1 + {{(1 - \beta )}^{t - 1}}} \right)\tau {\rho _0} + G + {\sigma ^2}} \right)\\&\quad \quad \cdot \left( {\left( {1 + {{(1 - \beta )}^{t - 1}}} \right){\rho _0} + \frac{{{\rho _0}}}{{\sqrt t }}} \right).
\end{aligned}
\end{equation} 

\end{theorem}
Theorem \ref{thm:sharpness} demonstrates that as $t$ increases, the difference between $S^{\text{EMA}}$ and $S^{\text{SAM}}$ gradually decreases. The perturbation estimated by the EMA can effectively approximate the original SAM perturbation.

\begin{table*}[ht]
    \centering
    \begin{tabular}{l|ccccccccc}
        \toprule
        RoBERTa    & \small{SST2} & \small{STS-B} & \small{RTE}  & \small{QQP}  & \small{QNLI} & \small{MRPC} & \small{MNLI} & \small{CoLA} & avg.$\uparrow$ \\
        \midrule
        FT$^\dagger$     & 96.4  & 92.4  & 86.6 & 92.2 & 94.7 & 90.9 & 90.2 & 68.0 & 88.9 \\
        \midrule 
        Adapter$^\dagger$      & \textbf{96.6}  & 91.9  & 80.1 & \textbf{91.7} & \textbf{94.8} & 89.7 & - & 67.8 & - \\
        LoRA$^*$             & 95.8  & 92.4  & 88.2 & 91.4 & 94.7 & 89.6 & {90.6} & 64.8 & 88.4 \\
        LoRA-oBAR$^*$  & \underline{96.0}  & \underline{92.6}  & 88.7 & \underline{91.6} & \textbf{94.8} & \underline{90.3} & {90.6} & 65.1 & {88.7}
        \\
        LoRA-nBAR$^*$        &  \underline{96.0}    & \underline{92.6}  & \underline{89.2}  & \underline{91.6}  & 94.7  & \underline{90.3}   &  \textbf{90.8}   &   \underline{65.6} & \underline{88.9}  \\
        %AdaLoRA$^\ddagger$        &  \underline{95.3}    & \underline{89.8}  & \underline{89.2}  & \underline{91.6}  & 94.7  & \underline{90.3}   &  \textbf{90.8}   &   \underline{65.6} & \underline{88.9}  \\
        EFMLoRA       &  \underline{96.3$_{\pm0.2}$}    & \textbf{92.7$_{\pm0.1}$}  & \textbf{89.3$_{\pm0.6}$}  &  \underline{91.6$_{\pm0.1}$} & \textbf{94.8$_{\pm0.1}$}  & \textbf{91.5$_{\pm0.4}$}   &  \underline{90.7$_{\pm0.1}$}   &   \textbf{68.0$_{\pm1.2}$} & \textbf{89.4}  \\
        \bottomrule
    \end{tabular}
    \caption{Experiments on finetuning RoBERTa (355M). Results marked with $\dagger$ are taken from \cite{hu2022lora}, and those with $*$ are taken from \cite{li2024implicit}.}
    \label{tab.lora-roberta-ft-full}
\end{table*}

\begin{table*}[ht]
    \centering
    \begin{tabular}{l|c|ccccc}
        \toprule
        \multirow{2}{*}{Model \& Method}         & \# Trainable & \multicolumn{5}{c}{E2E NLG Challenge} \\         & Parameters    & BLEU$\uparrow$       & NIST$\uparrow$ & MET$\uparrow$ & ROUGE-L$\uparrow$ & CIDEr$\uparrow$  \\
        \midrule
        GPT-2 M (FT)$^\dagger$   &  354.92M              & 68.2         & 8.62    & 46.2 & 71.0    & 2.47     \\\hline
        GPT-2 M (Adapter$^\textbf{L}$)$^\dagger$  & 0.37M                & 66.3         & 8.41    & 45.0 & 69.8    & 2.40     \\
        GPT-2 M (LoRA)    &  0.35M             & 69.2         & \underline{8.72}    & \underline{46.5} & \underline{71.5}    & \underline{2.51}     \\
        GPT-2 M (FMLoRA)    &  0.35M             & \underline{69.2}         & \underline{8.72}    & \bf{46.6} & \underline{71.5}    & \underline{2.51}     \\
        GPT-2 M (EFMLoRA)     &   0.35M           & \bf{69.7}         & \bf{8.77}    & \bf{46.6} & \bf{71.7}    & \bf{2.53}     \\ \midrule
        GPT-2 L (FT)$^\dagger$    &  774.03M             & 68.5         & 8.78    & 46.0 & 69.9    & 2.45     \\\hline
        GPT-2 L (Adapter$^\textbf{L}$)$^\dagger$   &   0.88M             & 69.1   & 8.68    & \underline{46.3} & 71.4   & 2.49     \\
        GPT-2 L (LoRA)    &     0.77M          & 69.9        & 8.82    & \bf{46.8} & \bf{71.8}    & \bf{2.53}     \\
        GPT-2 L (FMLoRA)    &     0.77M          & \underline{70.0}        & \underline{8.83}    & \bf{46.8} & \bf{71.8}    & \bf{2.53}     \\
        GPT-2 L (EFMLoRA)  &    0.77M             & \bf{70.2}         & \bf{8.84}   & \bf{46.8} & \bf{71.8}   & \underline{2.52}     \\
        \bottomrule
    \end{tabular}
    \caption{GPT-2 medium (M) and large (L) with different adaptation methods on the E2E NLG Challenge. Results marked with $\dagger$ are taken from \cite{hu2022lora}.}
    \label{tab.lora-gpt2}
\end{table*}

\subsection{Memory and Time Complexity}
LoRA reduces the number of trainable parameters by decomposing weight updates as $\Delta \mathbf{W} \approx \mathbf{B}\mathbf{A}$, where $\mathbf{B} \in \mathbb{R}^{n \times r}$ and $\mathbf{A} \in \mathbb{R}^{r \times m}$ with $r \ll \min(n, m)$. Both FMLoRA and EFMLoRA retain this parameter efficiency:
\begin{equation}\label{}
\begin{aligned}
\text{P}_{\text{LoRA}} &= \text{P}_{\text{FMLoRA}} = \text{P}_{\text{EFMLoRA}}\\&= O(nr+rm) \ll O(nm). 
\end{aligned}
\end{equation}
However, FMLoRA and EFMLoRA introduce additional memory overhead. Specifically, FMLoRA temporarily stores the original values of $\mathbf{B}$ and $\mathbf{A}$, as well as the gradients of $\mathbf{A}$. The memory usage of FMLoRA is calibrated as follows:
\begin{equation}\label{}
\begin{aligned}
\text{M}_{\text{FMLoRA}} = \text{M}_{\text{LoRA}} + O(1.5\times(nr+rm)), 
\end{aligned}
\end{equation}
where $\text{M}_{\text{LoRA}}$ indicates the memory required by LoRA. The memory of EFMLoRA needs to maintain the EMA perturbation on $\mathbf{B}$ as follows:
\begin{equation}\label{}
\begin{aligned}
\text{M}_{\text{EFMLoRA}} = \text{M}_{\text{LoRA}} + O(2\times(nr+rm)). 
\end{aligned}
\end{equation}
Notably, modern optimizers like AdamW already require $O(2\times(nr+rm))$ memory for momentum and second-moment statistics when applied to LoRA.

For time complexity, suppose that the time complexity of optimizing the model with LoRA is $O(T)$, which mainly includes the time for forward and backward. Theoretically, the time complexity of FMLoRA is approximately as follows:
\begin{equation}\label{}
\begin{aligned}
\text{T}_{\text{FMLoRA}} \approx O(2T) = 2 \times \text{T}_{\text{LoRA}}. 
\end{aligned}
\end{equation}
In contrast, the time complexity of EFMLoRA can be approximated as follows:
\begin{equation}\label{}
\begin{aligned}
\text{T}_{\text{EFMLoRA}} \approx O(T) =  \text{T}_{\text{LoRA}}. 
\end{aligned}
\end{equation}
We implement QR decomposition by Householder transformations, with time complexity of $O (r^2n)$ for an $r\times n$ matrix, e.g., $r$ is rank, $n$ is the input dimension in LORA. 

\section{Experiments and Discussions}
The best and second-best results are highlighted in bold and underline, respectively. Additional experimental details are provided in the supplementary file.

\subsection{Experiments on Large Language Models}
\textbf{Few-shot with RoBERTa-large.} We first consider few-shot learning with EFMLoRA. Following the setup of \cite{li2024implicit}, we adopt RoBERTa-large—a 355M-parameter language model—as the backbone. The results in Table \ref{fs_roberta} show that FMLoRA outperforms all other methods with the highest average score (83.1), particularly excelling on SST-2, SNLI, and MNLI. EFMLoRA follows closely with an average score of 82.3. It consistently surpasses baseline LoRA ($+2.3$), LoRA-SAM ($+1.0$), and both BAR variants. These results highlight its superior generalization ability under distribution shift and limited supervision. We conjecture that the performance gap between SAM and EFMLoRA comes from EFMLoRA eliminating the mutual interference between perturbations in the two low-rank subspaces.

\textbf{Fine-tuning with RoBERTa-large.} We apply EFMLoRA to finetune RoBERTa-large. Our implementation follows \cite{hu2022lora}, using the same hyperparameters as those in its GitHub repository. %We report results on the development set, as in previous work \cite{hu2022lora}, since the test set is not publicly available. 
The results can be found in Table \ref{tab.lora-roberta-ft-full}. we observe that EFMLoRA achieves the highest scores on all datasets, and achieves the highest accuracy on average over these datasets. Specifically, on average over these datasets, EFMLoRA surpasses standard LoRA with a margin of $1.0$. 
Additionally, EFMLoRA even achieve better performance than full fine-tuning on some datasets. This superior performance may be attributed to overfitting in full fine-tuning, where optimizing all model parameters can lead to overfitting on the training data, thus reducing the model’s generalization to the test set. This effect is particularly pronounced on small datasets, such as MRPC, which contains only 3.7k training data.

\textbf{Fine-tuning with GPT-2.} Having shown that FMLoRA is effective for NLU tasks, we now explore whether EFMLoRA can improve LoRA in NLG models like GPT-2 Medium and Large \cite{radford2019language}. To enable a direct comparison, we adopt the experimental setup of \cite{li2021prefix} with minimal deviation. Table~\ref{tab.lora-gpt2} demonstrates the effectiveness of EFMLoRA on the E2E NLG Challenge \cite{novikova2017e2e} with GPT-2 Medium and Large models. Compared with existing PEFT methods such as Adapter and LoRA, EFMLoRA consistently achieves superior performance across all metrics. Notably, it achieves this improvement without increasing the number of trainable parameters, maintaining the same efficiency as standard LoRA.

\subsection{Experiments on Vision Language Models}
\textbf{Few-shot with CLIP.} Recent advances in few-shot adaptation of Vision-Language Models (VLMs) have significantly enhanced their generalization. CLIP-LoRA \cite{zanella2024low} explores the application of LoRA in this few-shot VLM setting. In our work, we also apply FMLoRA and EFMLoRA to VLMs to evaluate their effectiveness. For a fair comparison, our experimental setup follows that of CLIP-LoRA. We
consider five datasets for fine-grained classification of
satellite imagery (EuroSAT \cite{helber2019eurosat}, Ox-fordPets \cite{parkhi2012cats}, Flower102 \cite{nilsback2008automated}, Caltech101 \cite{fei2004learning}, DTD \cite{cimpoi2014describing}). These datasets offer a thorough benchmarking framework for evaluating few-shot visual classification tasks. Table~\ref{tab.clip} demonstrates that FMLoRA and EFMLoRA outperformed Adapter and LoRA in most settings. In the low-data regimes (1-shot and 4-shot), EFMLoRA shows clear advantages. These results highlight the effectiveness of EFMLoRA in improving generalization in few-shot adaptation of vision-language models.

\begin{table}[ht]
    \centering
    \setlength{\tabcolsep}{1mm}
    \begin{tabular}{c|c|ccccc}
        \toprule
        Shots               & Method       & Eur.$\uparrow$ & Pets$\uparrow$ & Flo.$\uparrow$ & Cal.$\uparrow$ & DTD$\uparrow$  \\
        \midrule
        0                   & CLIP         & 47.5    & 89.1 & 71.4    & 92.9    & 43.6 \\
        \midrule 
        \multirow{4}{*}{1}  & Adapter & 49.3    & 89.0 & 71.3    & 92.0    & 44.2 \\
        & LoRA    & {72.3}    & \underline{92.3} & \bf{83.2}    & {93.7}    & {54.3} \\
                    & FMLoRA    & \underline{72.6}    & \bf{92.8} & \underline{82.8}    & \bf{94.5}    & \bf{54.9} \\
        & EFMLoRA    & \bf{78.3}    & \bf{92.8} & {81.0}    & \underline{93.9}    & \underline{54.6} \\ \midrule
\multirow{4}{*}{4}  & Adapter & 51.2    & 90.8 & 73.1    & 94.0    & 46.1 \\
                    & LoRA    & {84.9}    & {91.0} & {93.7}    & \underline{95.2}    & {63.8} \\
                    & FMLoRA    & \bf{90.0}      & \bf{93.1} & \bf{94.9}    & \bf{95.6}    & \bf{65.7} \\ 
                    & EFMLoRA    & \underline{87.6}    & \underline{91.1} & \underline{94.0}    & \bf{95.6}    & \underline{65.0} \\ \midrule
\multirow{4}{*}{16} & Adapter & 71.4    & 92.3 & 92.9    & 94.9    & 59.4 \\
                    & LoRA    & \underline{92.1}    & \underline{92.4} & {98.0}    & {96.4}    & \underline{72.0} \\
                    & FMLoRA    & \bf{92.2}    & \bf{93.4} & \bf{98.5}    & \underline{96.5}    & \bf{72.7} \\
                    & EFMLoRA    & {91.6}    & {91.5} & \underline{98.1}    & \bf{96.6}    & {71.9} \\ 
        \bottomrule
    \end{tabular}
    \caption{Detailed results for five datasets with CLIP-Adapter, CLIP-LoRA and EFMLoRA.}
    \label{tab.clip}
\end{table}

\textbf{Fine-tuning with Qwen-VL-Chat.} Qwen-VL-Chat \cite{Bai2023QwenVLAV} is a multimodal conversational large language model capable of understanding both images and text. We apply EFMLoRA to fine-tune Qwen-VL-Chat, following the same experimental setup as in \cite{zhou2024empirical}. Table \ref{tab.Qwen} presents the results on the ScienceQA \cite{lu2022learn} and VizWiz \cite{gurari2018vizwiz} datasets. The results in Table \ref{tab.Qwen} demonstrate that the perturbation size $\rho$ significantly influences the performance of EFMLoRA when fine-tuning Qwen-VL-Chat. By tuning $\rho$, EFMLoRA adapts to different tasks, enabling improved generalization—achieving higher accuracy than LoRA. Specifically, a larger $\rho$ (e.g., $\rho=0.2$) yields the best accuracy on ScienceQA, while a smaller $\rho$ (e.g., $\rho=0.05$) performs better on VizWiz. This suggests that different tasks benefit from different levels of perturbation. Therefore, selecting an appropriate $\rho$ based on the task characteristics is crucial for achieving optimal fine-tuning performance on multimodal large language models.

\begin{table}[t]
    \centering
    \setlength{\tabcolsep}{1mm}
    \begin{tabular}{l|c|cc}
\toprule
Method                                                                             & $\rho$ & SQA$\uparrow$  & VizWiz$\uparrow$ \\ \midrule
\begin{tabular}[c]{@{}l@{}}LoRA\end{tabular}                        & -    & \underline{90.1} & 50.69  \\ \midrule
\multirow{4}{*}{\begin{tabular}[c]{@{}l@{}}EFMLoRA\end{tabular}} 
& 0.05 & {90.0} & \bf{51.7}  \\
& 0.1 & {90.0} & {50.6}  \\
& 0.2 & \bf{91.6} & \underline{51.0}  \\
 & 0.6  & 89.6 & 50.0  \\
  %& 1   & \bf{90.3} & 49.38  \\
  \bottomrule
\end{tabular}
    \caption{EFMLoRA Fine-Tuning Results on Qwen-VL-Chat with different $\rho$.}
    \label{tab.Qwen}
\end{table}

\begin{table}[t]
\centering
\setlength{\tabcolsep}{1mm}
\begin{tabular}{l|cccc}
\toprule
\multirow{2}{*}{Methods} & \multicolumn{2}{c}{GPT-2 Medium} & \multicolumn{2}{c}{GPT-2 Large} \\  
                          & Memory$\downarrow$      & Time$\downarrow$     & Memory$\downarrow$          & Time$\downarrow$         \\ \midrule
LoRA   &  \bf{23.6}  & \bf{4.30}    & \bf{23.2} & \bf{8.45}     \\
FMLoRA & \underline{24.0} & 9.10  & \underline{23.6} & 17.47  \\
EFMLoRA& \underline{24.0} & \underline{4.80}   & \bf{23.2} & \underline{10.00}  \\ \bottomrule
\end{tabular}
\caption{Runtime (Hour) and memory (GB) of LoRA, FMLoRA and EFMLoRA on fine-tuning GPT-2 Medium/Large.
}
\label{time_fs_gpt2} 
\end{table}

\subsection{Runtime and Memory Consumption}
The results in Table~\ref{time_fs_gpt2} confirm the theoretical time complexity analysis. As expected, FMLoRA has approximately double the runtime of LoRA (2.1× on both GPT-2 Medium and Large), consistent with its theoretical complexity of $O(2T)$ due to two forward and backward passes for sharpness optimization. In contrast, EFMLoRA operates with near-LoRA efficiency, requiring only 1.1× and 1.2× more time on GPT-2 Medium and Large, respectively. This supports the theoretical claim that EFMLoRA maintains a time complexity close to $O(T)$ while benefiting from sharpness-aware optimization. In addition, EFMLoRA maintains a memory usage almost identical to that of LoRA, with only negligible increases (less than 0.4 GB across both model scales). 
These results demonstrate that EFMLoRA achieves near-LoRA efficiency in both memory and runtime.

\subsection{Conclusion}
In this work, we propose FMLoRA, a novel PEFT method that integrates sharpness-aware optimization into the LoRA framework to promote convergence toward flatter minima. We theoretically demonstrate that perturbations in the full parameter space can be equivalently represented within the low-rank subspace. To improve computational efficiency, we introduce EFMLoRA, which leverages an exponential moving average to approximate perturbations, significantly reducing runtime overhead while maintaining effectiveness. Extensive experiments across various large language and vision-language models demonstrate that EFMLoRA achieves comparable or even superior generalization performance to full fine-tuning and LoRA. Our results emphasize the importance of reducing sharpness to improve generalization in PEFT methods, offering valuable insights and practical tools for future research on the link between sharpness and generalization in LLMs and beyond.

\section{A. Proofs}
\subsection{A.1 Proof of Eq.(10) and Eq.(11)}
\begin{proof}
we propose to approximate the unknown gradient $\nabla L_\mathbf{W}(\mathbf{W})$ using standard LoRA gradients, which can be computed in two ways:
\begin{align}
(1) \nabla L_\mathbf{B}&(\mathbf{W_0}+ s\mathbf{BA}) = s\nabla L_\mathbf{W}(\mathbf{W})\mathbf{A}^\top \nonumber \\&\Rightarrow  
\quad \nabla L_\mathbf{W}(\mathbf{W})= \frac{1}{s}  \nabla L_\mathbf{B}(\mathbf{W_0}+ s\mathbf{BA})(\mathbf{A}^\top)^+, \\
(2) \nabla L_\mathbf{A}&(\mathbf{W_0}+ s\mathbf{BA}) = s\mathbf{B}^\top\nabla L_\mathbf{W}(\mathbf{W}) \nonumber \\ & \Rightarrow 
\quad \nabla L_\mathbf{W}(\mathbf{W})=\frac{1}{s} (\mathbf{B}^\top)^+\nabla L_\mathbf{A}(\mathbf{W_0}+ s\mathbf{BA}),
\end{align}
\end{proof}

\subsection{A.2 Proof of Theorem 1}

\begin{proof}
The update process of the FMLoRA is as follows:
\begin{equation}\label{equ:}
\begin{aligned}
{\tilde{\mathbf{x}}_t} = {{\mathbf{x}}_t} + \rho \frac{1}{s}\frac{{{\mathbf{G}_t}}}{{\left\| {{\mathbf{G}_t}} \right\|}_F}{\mathbf{y}_t}^+&, \quad {\tilde{\mathbf{y}}_t} = {{\mathbf{y}}_t}\\
{\mathbf{g}_{{\tilde{\mathbf{x}}_t}}} = {{\tilde{\mathbf{G}}}_t}\tilde{\mathbf{y}}_t&, \quad{\mathbf{g}_{{\tilde{\mathbf{y}}_t}}} = {{\tilde{\mathbf{G}}}_t}^{\top}\tilde{\mathbf{x}}_t\\
{{\mathbf{x}}_{t + 1}} = {{\mathbf{x}}_t} - \eta {\mathbf{g}_{{\tilde{\mathbf{x}}_t}}}&, \quad{{\mathbf{y}}_{t + 1}} = {{\mathbf{y}}_t} - \eta {\mathbf{g}_{{\tilde{\mathbf{y}}_t}}}
\end{aligned}
\end{equation} 
where ${\mathbf{x}}_t=\text{Vector}(\mathbf{B}_t)$ is the vectorized form of matrix $\mathbf{B}_t$, ${\mathbf{y}}_t$ is the vectorized form of matrix $\mathbf{A}_t$, ${\mathbf{G}_t}=\nabla L({\mathbf{x}}_t{\mathbf{y}}_t^{\top})$ is the gradient of the full parameter space at the original point during gradient descent, ${\tilde{\mathbf{G}}_t}=\nabla L(\tilde{\mathbf{x}}_t\tilde{\mathbf{y}}_t^{\top})$ is the gradient of the full parameter space at the perturbed point, and $\mathbf{y}^+$ is the pseudo inverse of $\mathbf{y}$. Let balancedness $B_t:=\frac{1}{2}(||\mathbf{x}_t||^2-||\mathbf{y}_t||^2)$. Then, we have that:
\begin{equation}\label{equ:}
\begin{aligned}
&\frac{1}{2}\frac{{d({{\left\| {{\mathbf{x}_t}} \right\|}^2} - {{\left\| {{\mathbf{y}_t}} \right\|}^2})}}{{dt}}\\
&= \frac{1}{2}\frac{{d({{\left\| {{\mathbf{x}_t}} \right\|}^2})}}{{dt}} - \frac{1}{2}\frac{{d({{\left\| {{\mathbf{y}_t}} \right\|}^2})}}{{dt}}\\
&= {\mathbf{x}_t}^\top\frac{{d{\mathbf{x}_t}}}{{dt}} - {{\mathbf{y}}_t}^\top\frac{{d{\mathbf{y}_t}}}{{dt}}\\
&= -{\mathbf{x}_t}^\top({\mathbf{\tilde G}_t}{\mathbf{y}_t}) + ({\mathbf{y}_t}^\top({\mathbf{\tilde G}_t}^\top({{\mathbf{x}}_t} + \rho \frac{1}{s}\frac{{{\mathbf{G}_{{t}}}}}{{{{\left\| {\mathbf{G}_t} \right\|}_F}}}\mathbf{y}_t^+ )))\\
&= -{\mathbf{x}_t}^\top({\mathbf{\tilde G}_t}{\mathbf{y}_t}) + ({\mathbf{y}_t}^\top({\mathbf{\tilde G}_t}^\top{\mathbf{{x}}_t} + \rho \frac{1}{s}{\mathbf{\tilde G}_t}^\top\frac{{{\mathbf{G}_{{t}}}}}{{{{\left\| {{\mathbf{G}_t}} \right\|}_F}}}\mathbf{y}_t^ + ))\\
&= -{\mathbf{x}_t}^\top{\mathbf{\tilde G}_t}{\mathbf{y}_t} + ({\mathbf{x}_t}^\top{\mathbf{\tilde G}_t}{\mathbf{y}_t}){^\top} + \rho \frac{1}{s}{\mathbf{y}_t}^\top{\mathbf{\tilde G}_t}^\top\frac{{{\mathbf{G}_{{t}}}}}{{{{\left\| {{\mathbf{G}_t}} \right\|}_F}}}\mathbf{y}_t^ + \\
&=  \rho \frac{1}{s}{\mathbf{y}_t}^\top{\mathbf{\tilde G}_t}^\top\frac{{{\mathbf{G}_{{t}}}}}{{{{\left\| {{\mathbf{G}_t}} \right\|}_F}}}\mathbf{y}_t^ + \\
&=  \rho \frac{1}{s}\frac{1}{{{{\left\| {{\mathbf{G}_t}} \right\|}_F}}}\left[ {{\mathbf{y}_t}^\top{\mathbf{\tilde G}_t}^\top{\mathbf{G}_{{t}}}\mathbf{y}_t^ + } \right]
\end{aligned}
\end{equation} 
Because $\frac{1}{{{s}}}\mathbf{g_x} = {\mathbf{G}_{{t}}}{\mathbf{y}_t}$ and ${\mathbf{g}_{{{{\mathbf{\tilde x}}}_t}}} = {\mathbf{\tilde G}_t}{\mathbf{\tilde y}_t}$, we have:
\begin{equation}\label{equ:}
\begin{aligned}
&\frac{1}{2}\frac{{d({{\left\| {{\mathbf{x}_t}} \right\|}^2} - {{\left\| {{\mathbf{y}_t}} \right\|}^2})}}{{dt}}\\
&=\rho \frac{1}{s}\frac{1}{{{{\left\| {{\mathbf{G}_t}} \right\|}_F}}}\left[ {{\mathbf{y}_t}^\top{\mathbf{\tilde G}_t}^\top{\mathbf{G}_{{t}}}\mathbf{y}_t^ + } \right]\\
&=  \rho \frac{1}{{{s^2}}}\frac{1}{{{{\left\| {{\mathbf{G}_t}} \right\|}_F}}}\left[ {{{({\mathbf{\tilde G}_t}{\mathbf{y}_t})}^\top}\mathbf{{g}_x}{{(\mathbf{y}_t^\top)}^ + }\mathbf{y}_t^ + } \right]\\
&=  \rho \frac{1}{{{s^2}}}\frac{1}{{{{\left\| {{\mathbf{G}_t}} \right\|}_F}}}\left[ {{{({\mathbf{\tilde G}_t}{{}\mathbf{y}_t})}^\top}\mathbf{g_x}{{(\mathbf{y}_t^ + )}^\top}\mathbf{y}_t^ + } \right]\\
&=  \rho \frac{1}{{{s^2}}}\frac{1}{{{{\left\| {{\mathbf{G}_t}} \right\|}_F}}}\left[ {{{({\mathbf{\tilde G}_t}{\mathbf{y}_t})}^\top}\mathbf{g_x}{{\left\| {\mathbf{y}_t^ + } \right\|}^2}} \right] \\
&=  \rho \frac{1}{{{s^2}}}\frac{1}{{{{\left\| {{\mathbf{G}_t}} \right\|}_F}}}\left[ {{{({\mathbf{g}_{{{\mathbf{{\tilde x}}}_t}}}({\mathbf{\tilde y}_t}^ + )^\top {\mathbf{y}_t})}^\top}\mathbf{g_x}{{\left\| {\mathbf{y}_t^ + } \right\|}^2}} \right]\\
&=  \rho \frac{1}{{{s^2}}}\frac{1}{{{{\left\| {{\mathbf{G}_t}} \right\|}_F}}}\left[ {{\mathbf{g}_{{{{\mathbf{\tilde x}}}_t}}}^\top\mathbf{g_x}{{\left\| {\mathbf{y}_t^ + } \right\|}^2}} \right]
\end{aligned}
\end{equation} 
Taking the absolute value of balancedness $B_t$ gives:
\begin{equation}\label{equ:}
\begin{aligned}
&\left| {\frac{1}{2}\frac{{d({{\left\| {{\mathbf{x}_t}} \right\|}^2} - {{\left\| {{\mathbf{y}_t}} \right\|}^2})}}{{dt}}} \right|\\
&= \left| {  \rho \frac{1}{s}\frac{{{{\left\| {\mathbf{y}_t^ + } \right\|}^2}}}{{{{\left\| {{\mathbf{{g}}_\mathbf{x}}{{(\mathbf{y}_t^+)}^ \top }} \right\|}_F}}}( {{\mathbf{g}_{{{{\mathbf{\tilde x}}}_t}}}^\top\mathbf{g_x}} )} \right| \\
&= \left| { \rho \frac{1}{s}\frac{{{{\left\| {\mathbf{y}_t^ + } \right\|}^2}}}{{\left\| {\mathbf{{g}_x}} \right\|\left\| {{{(\mathbf{y}_t^ + )}^\top}} \right\|}}( {{\mathbf{g}_{{{{\mathbf{\tilde x}}}_t}}}^\top\mathbf{g_x}} )} \right|\\
&= \left| {\rho \frac{1}{s}\frac{{\left\| {\mathbf{y}_t^ + } \right\|}}{{\left\| {\mathbf{{g}_x}} \right\|}}( {{\mathbf{g}_{{{{\mathbf{\tilde x}}}_t}}}^\top\mathbf{g_x}} )} \right|\\
&\le \left| {\rho \frac{1}{s}\frac{{\left\| {\mathbf{y}_t^ + } \right\|}}{{\left\| {\mathbf{g_x}} \right\|}}\left\| {{\mathbf{g}_{{{{\mathbf{\tilde x}}}_t}}}} \right\|\left\| {\mathbf{{{g}}_x}} \right\|} \right|\\
&= \left| {\rho \frac{1}{s}\left\| {\mathbf{y}_t^ + } \right\|\left\| {{\mathbf{g}_{{{{\mathbf{\tilde x}}}_t}}}} \right\|} \right|\\
&= \left| {\rho \frac{1}{s}\left\| {\frac{{\mathbf{y}_t^{\top}}}{{{{\left\| {\mathbf{y}_t} \right\|}^2}}}} \right\|\left\| {{\mathbf{g}_{{{{\mathbf{\tilde x}}}_t}}}} \right\|} \right|\\
%&= \left| {\rho \frac{1}{s}\frac{{\left\| {y_t^\top} \right\|}}{{{{\left\| {y_t^{}} \right\|}^2}}}\left\| {{g_{{{{\rm{\tilde x}}}_t}}}} \right\|} \right|\\
&= \left| {\rho \frac{1}{s}\frac{1}{{\left\| {\mathbf{y}_t} \right\|}}\left\| {{\mathbf{g}_{{{{\mathbf{\tilde x}}}_t}}}} \right\|} \right|
\end{aligned}
\end{equation} 
The proof is thus completed.
\end{proof}

\begin{lemma}
Let $A_{t+1} = \alpha A_t + \beta$ with some $\alpha \in (0, 1)$, then we have
\[
A_{t+1} \leq \alpha^{t+1} A_0 + \frac{\beta}{1 - \alpha}.
\]
\end{lemma}

\begin{proof}
The proof can be completed by simply unrolling $A_{t+1}$ and using the fact 
$1 + \alpha + \alpha^2 + \dots + \alpha^t \leq \frac{1}{1 - \alpha}$.
\end{proof}

\subsection{A.3 Proof of Theorem 2}

\begin{proof}
Assume that $\bm{\varepsilon}_t$ is the perturbation at time step $t$, and $\bm{\hat{\varepsilon}}_{t-1}$ is the EMA perturbation from the previous step. Let $\nabla L(\mathbf{w}_t + \bm{\hat{\varepsilon}}_{t-1})$ denote the gradient used for updating at time $t$. The standard SAM perturbation at step $t$ is defined as
$\bm{\tilde{\varepsilon}}_t = \rho_t \frac{\nabla L(\mathbf{w}_t)}{\|\nabla L(\mathbf{w}_t)\|}$,
and the EMA perturbation at step $t$ is computed as $\bm{\hat{\varepsilon}}_t = (1 - \beta) \bm{\hat{\varepsilon}}_{t-1} + \beta \bm{\varepsilon}_t$.
Based on Assumption 4, we have that:
\begin{align}
&\left[ {L({\mathbf{w}_t} + {\bm{\hat{\varepsilon} }_{t-1}}) - L({\mathbf{w}_t})} \right] - \left[ {L({\mathbf{w}_t} + {\bm{\tilde \varepsilon }_t}) - L({\mathbf{w}_t})} \right]\\ \nonumber
&= L({\mathbf{w}_t} + {\bm{\hat \varepsilon }_{t-1}}) - L({\mathbf{w}_t} + {\bm{\tilde \varepsilon }_{t}})\\ \nonumber
&\le  - \nabla L{({\mathbf{w}_t} + {\bm{\hat \varepsilon }_{t-1}})^\top}({\mathbf{w}_t} + {\bm{\tilde \varepsilon }_t} - {\mathbf{w}_t} - {\bm{\hat \varepsilon }_{t-1}})\\ \nonumber
&= \nabla L{({\mathbf{w}_t} + {\bm{\hat \varepsilon }_{t-1}})^\top}({\bm{\hat \varepsilon }_{t-1}} - {\bm{\tilde \varepsilon }_t})\\ \nonumber
&\le \left| {\nabla L{{({\mathbf{w}_t} + {\bm{\hat \varepsilon }_{t-1}})}^\top}({\bm{\hat \varepsilon }_{t-1}} - {\bm{\tilde \varepsilon }_t})} \right|\\
&\le \left\| {\nabla L({\mathbf{w}_t} + {\bm{\hat \varepsilon }_{t-1}})} \right\|\left\| {{\bm{\hat \varepsilon }_{t-1}} - {\bm{\tilde \varepsilon }_t}} \right\| \label{equ:T2_1}
\end{align}
For the first term $\left\| {\nabla L({\mathbf{w}_t} + {\bm{\hat \varepsilon }_{t-1}})} \right\|$ in Eq.~\eqref{equ:T2_1}, Based on Assumption 1, Assumption 2 and Lemma 1, we have:
\begin{equation}\label{equ:}
\begin{aligned}
&\left\| {\nabla L({\mathbf{w}_t} + {\bm{\hat \varepsilon }_{t-1}})} \right\|\\
&= \left\| {\nabla L({\mathbf{w}_t} + {\bm{\hat \varepsilon }_{t-1}}) - \nabla L({\mathbf{w}_t}) + \nabla L({\mathbf{w}_t})} \right\|\\
&\le \left\| {\nabla L({\mathbf{w}_t} + {\bm{\hat \varepsilon }_{t-1}}) - \nabla L({\mathbf{w}_t})} \right\| + \left\| {\nabla L({\mathbf{w}_t})} \right\|\\
&\le \tau \left\| {{\mathbf{w}_t} + {\bm{\hat \varepsilon }_{t-1}} - {\mathbf{w}_t}} \right\| + \left\| {\nabla L({\mathbf{w}_t})} \right\|\\
&= \tau \left\| {{\bm{\hat \varepsilon }_{t-1}}} \right\| + \left\| {\nabla L({\mathbf{w}_t}) - \nabla {L_{\rm{D}}}({\mathbf{w}_t}) + \nabla {L_{\rm{D}}}({\mathbf{w}_t})} \right\|\\
&= \tau \left\| {{\bm{\hat \varepsilon }_{t-1}}} \right\| + \left\| {\nabla {L_{\rm{D}}}({\mathbf{w}_t})} \right\| + {\sigma ^2}\\
&= \tau \left\| {(1 - \beta ){\bm{\hat \varepsilon }_{t - 2}} + \beta \bm{\varepsilon}_{t-1}} \right\| + G + {\sigma ^2}\\
&\le \tau ((1 - \beta )\left\| {{\bm{\hat \varepsilon }_{t - 2}}} \right\| + \beta {\rho_0})+ G + {\sigma ^2}\\
&\le \tau {(1 - \beta )^{t-1}}\left\| {{\bm{\hat \varepsilon }_0}} \right\| + \tau {\rho_0} + G + {\sigma ^2}\\
\end{aligned}
\end{equation} 
For the second term $\left\| {{\bm{\hat \varepsilon }_{t-1}} - {\bm{\tilde \varepsilon }_t}} \right\|$ in Eq.~\eqref{equ:T2_1}, we have:
\begin{equation}\label{equ:}
\begin{aligned}
&\left\| {{\bm{\hat \varepsilon }_{t-1}} - {\bm{\tilde \varepsilon }_t}} \right\|\\
&\le \left\| {{\bm{\hat \varepsilon }_{t-1}}} \right\| + \left\| {{\bm{\tilde \varepsilon }_t}} \right\|\\
&= \left\| {{\bm{\hat \varepsilon }_{t-1}}} \right\| + {\rho _{\rm{t}}}\\
%&= \left\| {(1 - \beta ){\bm{\hat \varepsilon }_{t - 1}} + \beta \varepsilon _t} \right\| + {\rho _t}\\
%&\le (1 - \beta )\left\| {{\bm{\hat \varepsilon }_{t - 1}}} \right\| + \beta {\rho _0} + {\rho _t}\\
&\le {(1 - \beta )^{t-1}}\left\| {{\bm{\hat \varepsilon }_0}} \right\| + {\rho _0} + {\rho _t}\\
% &= {\rho _0}{(1 - \beta )^{t}} + {\rho _0} + {\rho _t}
\end{aligned}
\end{equation} 
Let ${\bm{\hat \varepsilon }_0} = {\bm{\tilde \varepsilon }_0}=\rho_0 \frac{\nabla L(\mathbf{w}_0)}{\|\nabla L(\mathbf{w}_0)\|}$, ${\rho _t} = \frac{{{\rho _0}}}{{\sqrt t }}$, we have:
\begin{equation}\label{equ:}
\begin{aligned}
&\left[ {L({\mathbf{w}_t} + {\bm{\hat \varepsilon }_{t-1}}) - L({\mathbf{w}_t})} \right] - \left[ {L({\mathbf{w}_t} + {\bm{\tilde \varepsilon }_t}) - L({\mathbf{w}_t})} \right]\\
&\le \left( \tau {(1 - \beta )^{t-1}}\left\| {{\bm{\hat \varepsilon }_0}} \right\| + \tau {\rho_0} + G + {\sigma ^2} \right)\\
&\quad \cdot \left( {(1 - \beta )^{t-1}}\left\| {{\bm{\hat \varepsilon }_0}} \right\| + {\rho _0} + {\rho _t} \right)\\
& = \left( {\left( {1 + {{(1 - \beta )}^{t - 1}}} \right)\tau {\rho _0} + G + {\sigma ^2}} \right)\\
& \quad\quad \cdot \left( {\left( {1 + {{(1 - \beta )}^{t - 1}}} \right){\rho _0} + \frac{{{\rho _0}}}{{\sqrt t }}} \right)\\
\end{aligned}
\end{equation} 
The proof is thus completed.
\end{proof}

\section{B. Experimental Details}
\subsection{B.1 Details on datasets}
Our evaluations are carried out on commonly-used datasets in the literature.

\textbf{Datasets for few-shot learning of RoBERTa-large.} We consider classification datasets: SST-2 \cite{socher2013recursive}, SST-5 \cite{socher2013recursive}, TREC \cite{voorhees2000building},
MNLI \cite{williams2018broad}, SNLI \cite{bowman2015large}, and RTE \cite{dagan2005pascal}. We follow Malladi et al. \cite{malladi2023kernel} in limiting the test set to 1, 000 examples for fast iteration. For training and validation, we set k = 512, which mean that we have 512 examples per class for both training and validation.

\begin{table*}[h]
    \footnotesize
    \addtolength{\tabcolsep}{-1pt}
    \centering
    \begin{tabular}{lcccccccc}
        \hline
        \toprule
        Dataset     & MNLI & SST-2 & MRPC & CoLA & QNLI & QQP & RTE & STS-B \\
        \midrule
                              Optimizer   & \multicolumn{8}{c}{AdamW} \\
                              Warmup Ratio & \multicolumn{8}{c}{0.06} \\
                              LR Schedule & \multicolumn{8}{c}{Linear} \\
        \midrule
        
                              Batch Size & 32 & 64 & 32 & 32 & 32 & 32 & 64 & 32 \\
                              Epochs & 10 & 10 & 20 & 20 & 10 & 20 & 20 & 10 \\
                              Learning Rate & 3E-04 & 4E-04 & 3E-04 & 3E-04 & 2E-04 & 3E-04 & 4E-04 & 3E-04 \\
                              LoRA Config. & \multicolumn{8}{c}{$r_q=r_v=8$} \\
                              LoRA $\alpha$ & \multicolumn{8}{c}{16} \\
                              $\rho$ for EFMLoRA  & \multicolumn{8}{c}{0.6} \\
                              $\beta$ for EFMLoRA  & \multicolumn{8}{c}{0.99} \\
                              scheduler for $\rho$  & \multicolumn{8}{c}{cosine} \\
                              Max Seq. Len. & 128 & 512 & 512 & 128 & 512 & 512 & 512 & 128 \\
        \bottomrule
    \end{tabular}
    \caption{The hyperparameters used for RoBERTa large with LoRA on the GLUE benchmark.}
    \label{hyp.ft_roberta}
\end{table*}

\begin{table}[t]
    \centering
    \begin{tabular}{lc}
        \toprule
        Hyper-parameters & Values \\
        \midrule
        LoRA $r$ (rank) & 8 \\
        LoRA $\alpha$  & 16 \\
        iterations & 1000 \\
        batchsize & 16 \\
        learning rate &   1$\times 10^{-4}$, 5$ \times 10^{-5}$ \\
        $\rho$ for FMLoRA & 0.3 \\
        $\beta$ for FMLoRA & 0.95 \\
        $\rho$ for EFMLoRA & 0.3 \\
        $\beta$ for EFMLoRA & 0.95 \\
        scheduler for $\rho$ & linear \\
        \bottomrule
    \end{tabular}
    \caption{Hyperparameters used for few-shot learning with RoBERTa-large.}
    \label{hyp.fs_roberta}
\end{table}

\begin{table}[t]
    \centering
    \begin{tabular}{lc}
        \toprule
        Hyper-parameters & Values \\
        \midrule
        LoRA $r$ (rank) & 4 \\
        LoRA $\alpha$  & 32 \\
        epochs & 5 \\
        batchsize & 8, 4 \\
        learning rate &   2$ \times 10^{-4}$ \\
        label Smooth & 0.1 \\
        $\rho$ for FMLoRA & 0.1\\
        $\beta$ for FMLoRA & 0.99 \\
        $\rho$ for EFMLoRA & 0.1\\
        $\beta$ for EFMLoRA & 0.99 \\
        scheduler for $\rho$ & cosine\\
        \midrule
        beam size & 10 \\
        length penalty  & 0.8 \\
        \bottomrule
    \end{tabular}
    \caption{Hyperparameters used for GPT2.}
    \label{tab.hyp_gpt2}
\end{table}

\textbf{GLUE benchmark.} GLUE is designed to provide a general-purpose evaluation of language understanding \citep{wangglue}. Those adopted in our work include MNLI (inference, \citep{williams2018broad}), SST-2 (sentiment analysis, \citep{socher2013recursive}), MRPC (paraphrase detection, \citep{dolan2005automatically}), CoLA (linguistic acceptability \citep{warstadt2019neural}), QNLI (inference \citep{rajpurkar2018know}), QQP\footnote{\url{https://quoradata.quora.com/First-Quora-Dataset-Release-Question-Pairs}} (question-answering), RTE\footnote{\url{https://paperswithcode.com/dataset/rte}} (inference), and STS-B (textual similarity \citep{cer2017semeval}). These datasets are released under different permissive licenses. 

\textbf{E2E NLG Challenge.} The E2E NLG Challenge dataset \cite{novikova2017e2e} is a standard benchmark for end-to-end data-to-text natural language generation. It consists of around 42,000 training instances, along with 4,600 each for validation and testing, all within the restaurant domain. Inputs are structured as sequences of slot-value pairs and paired with one or more reference texts. The dataset is released under the Creative Commons BY-NC-SA 4.0 license.

\textbf{Datasets for few-shot learning of CLIP.} We consider five datasets for fine-grained classification of satellite imagery (EuroSAT \cite{helber2019eurosat}), pet breeds (Ox-fordPets \cite{parkhi2012cats}), flowers (Flower102 \cite{nilsback2008automated}), general objects (Caltech101 \cite{fei2004learning}), textures (DTD \cite{cimpoi2014describing}). These datasets offer a thorough benchmarking framework for evaluating few-shot visual classification tasks.

\textbf{Datasets for fine-tuning with Qwen-VL-Chat.} We use two representative datasets: ScienceQA \cite{lu2022learn} and VizWiz \cite{gurari2018vizwiz}. ScienceQA is a multimodal multiple-choice QA dataset covering elementary science, with questions accompanied by text and images. VizWiz is a real-world visual QA dataset collected from blind users, featuring diverse and often low-quality images, posing challenges for robust multimodal understanding.

\subsection{B.2 Details on models}

We summarize the adopted language models in our evaluation. All model checkpoints are obtained from HuggingFace.

\textbf{RoBERTa-large.} This is a $355$M parameter model. The model checkpoint\footnote{\url{https://huggingface.co/FacebookAI/roberta-large}} is released under the MIT license.

\textbf{GPT2-medium.} This is a $345$M parameter model. Its checkpoint\footnote{\url{https://s3.amazonaws.com/models.huggingface.co/bert/gpt2-medium-pytorch_model.bin}} is under MIT License.

\textbf{GPT2-large.} This is a $774$M parameter model. Its checkpoint\footnote{\url{https://s3.amazonaws.com/models.huggingface.co/bert/gpt2-large-pytorch_model.bin}} is under MIT License.

\textbf{CLIP.} This is a model that learns to connect images and text by mapping them into a shared semantic space using contrastive learning. 

\textbf{Qwen-VL-Chat.} Qwen-VL-Chat \cite{Bai2023QwenVLAV} is a multimodal conversational large language model capable of understanding both images and text.

\begin{table}[t]
    \centering
    \begin{tabular}{lc}
\toprule
Hyper-parameters & Values   \\ \midrule
shots            & 1,4,16   \\
backbone         & ViT-B/16 \\
learning rate    & 2e-4     \\
batchsize        & 32       \\
LoRA $r$ (rank) & 2 \\
LoRA $\alpha$   & 1 \\
$\rho$ for FMLoRA   & 0.6      \\
$\beta$ for FMLoRA   & 0.99     \\ 
$\rho$ for EFMLoRA   & 0.1, 0.2, 0.5    \\
$\beta$ for EFMLoRA   & 0.99     \\ 
scheduler for $\rho$ & cosine\\\bottomrule
\end{tabular}
    \caption{Hyperparameters used for few-shot learning with CLIP.}
    \label{hyp.fs_CLIP}
\end{table}

\begin{algorithm}[t]
    %\begin{small}
    \caption{{Pseudocode of the FMLoRA}}
    {\bf Require:}
    The training dataset, the learning rate $\eta$, the batch size $b$, parameters $\rho$ and $\beta$.
    \begin{algorithmic}[1]
    \FOR{$t = 1,2,\cdot\cdot\cdot$}
    \STATE Randomly sample a mini-batch;
    \STATE Evaluate the gradient at the current point;
    \STATE Apply Equation (12) to compute the gradient in the full parameter space ${\Bar{\mathbf{g}}^\mathbf{W}}$;
    \STATE Use Equation (13) to calculate the perturbation $\Bar{\mathbf{E}}^\mathbf{W}$;
    \STATE Compute the perturbation $\Bar{\mathbf{E}}^\mathbf{B}=\frac{1}{s}\Bar{\mathbf{E}}^\mathbf{W}\mathbf{A}^+$ on matrix $\mathbf{B}$ according to Equation (14);
    \STATE Evaluate the gradient at the perturbed point ($\mathbf{W}_0$,  $\mathbf{B}+\Bar{\mathbf{E}}^\mathbf{B}$, $\mathbf{A}$);
    \STATE Return to the original (unperturbed) parameter point ($\mathbf{W}_0$, $\mathbf{B}$, $\mathbf{A}$);
    \STATE Update the weights using the gradient obtained in Step 6; 
    \ENDFOR
    \end{algorithmic}
    \label{algorithm:1}
   % \end{small}
\end{algorithm}

\subsection{B.3 Details on hyperparameters}
\textbf{Few-shot Learning with RoBERTa.}
We adopt the $k$-shot learning setup from \citep{malladi2023fine}, focusing on classification tasks with $k=512$ training samples per class and 1000 samples for testing. Prompt-based finetuning is used, following the same prompt templates as in \citep[Table 13]{malladi2023fine}. We use AdamW as the optimizer and tune hyperparameters based on Table \ref{hyp.fs_roberta}. All results are averaged over three random seeds.

\textbf{Fine-tuning with RoBERTa-large.}
AdamW is adopted as the base optimizer, and hyperparameters are in Table \ref{hyp.ft_roberta}. However, we employ single GPU rather than multiple ones and use gradient accumulation rather than parallelism due to memory constraint. We consider the GLUE benchmark and report the mismatched accuracy for MNLI, Matthew’s correlation for CoLA, Pearson correlation for STS-B, and accuracy for other datasets. Larger values indicate better results for all datasets. Experiments are conducted over three random trials for all datasets. 

\textbf{GPT2 medium/large on E2E NLG Challenge.}
We use the batch size, learning rate, and beam search beam size described
in \cite{hu2022lora}. AdamW is adopted as base optimizer.
The hyperparameters can be found in Table \ref{tab.hyp_gpt2}. The result for each run is taken from the last epoch.

\textbf{Few-shot Learning with CLIP.}
We follow the setting of previous work \cite{zanella2024low}. The hyperparameters are tuned from those in Table \ref{hyp.fs_CLIP}. We only apply low-rank matrices on the query, key and value matrices with $r = 2$. We regularize the input of the LoRA module by a dropout layer with $p = 0.25$. The number of iterations is set equal to 500 times N/K (the number of labeled samples per class). 

\textbf{Fine-tuning with Qwen-VL-Chat.}
We conduct experiments follow the setting of previous work \cite{zhou2024empirical}. The hyperparameters can be found in Table \ref{hyp.fs_Qwen}.

\begin{table}[t]
    \centering
    \begin{tabular}{lc}
\toprule
Hyper-parameters & Values   \\ \midrule
ViT            & Qwen-7B   \\
LLM         & ViT-G/16 \\
Connector    & CrossAttn     \\
Learning rate & 1e-5     \\
Learning rate schedule & cosine decay \\
Warm-up ratio & 0.01 \\
Weight decay  & 0.1 \\
Global batch size        & 128       \\
Epoch   & 3 \\
LoRA $r$ (rank) & 128 \\
$\rho$ for EFMLoRA   & 0.2,0.6,1      \\
$\beta$ for EFMLoRA   & 0.99     \\ 
scheduler for $\rho$ & cosine\\\bottomrule
\end{tabular}
    \caption{Hyperparameters used for fine-tuning with Qwen-VL-Chat.}
    \label{hyp.fs_Qwen}
\end{table}

\begin{figure}[t]
\centering
\includegraphics[width=0.7\columnwidth]{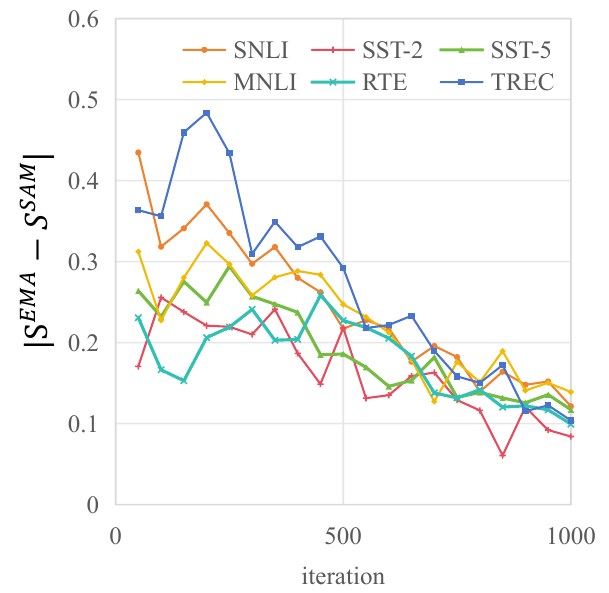} 
\caption{Approximation ability of EMA perturbations across datasets}
\label{fig_sema}
\end{figure}

\begin{figure*}[t]
\centering
\includegraphics[width=1.5\columnwidth]{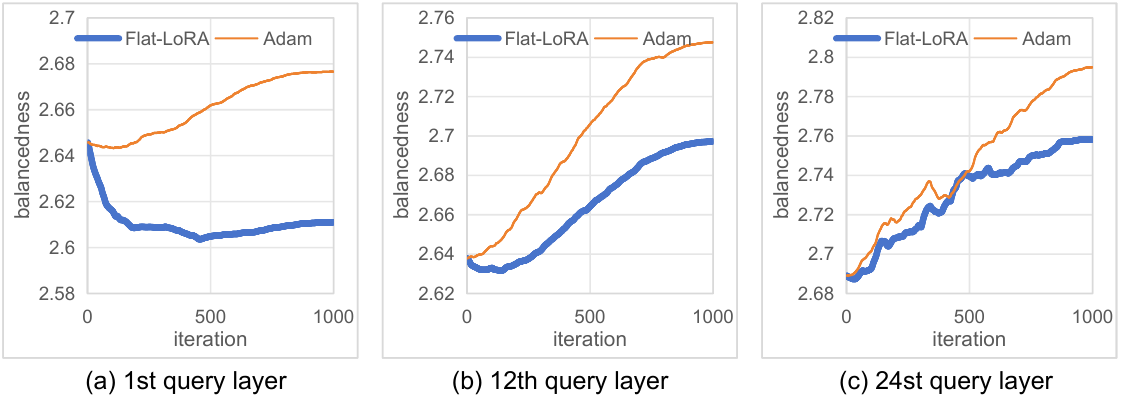} 
\caption{Evolution of balancedness across layers during training with Adam and FMLoRA.}
\label{fig_banan}
\end{figure*}

\section{C. Algorithm}
The two algorithms presented in \ref{algorithm:1} and \ref{algorithm:2} describe the training procedures of the proposed FMLoRA and its accelerated variant EFMLoRA. 

\begin{algorithm}[t]
    %\begin{small}
    \caption{{Pseudocode of the EFMLoRA}}
    {\bf Require:}
    The training dataset, the learning rate $\eta$, the batch size $b$, parameters $\rho$ and $\beta$.
    \begin{algorithmic}[1]
    \FOR{$t = 1,2,\cdot\cdot\cdot$}
    \STATE Randomly sample a mini-batch;
    \IF{$t = 1$}
    \STATE Evaluate the gradient at the current point;
    %\STATE Apply Equation (12) to compute the gradient in the full parameter space ${\Bar{\mathbf{g}}^\mathbf{W}}$;
    %\STATE Use Equation (13) to calculate the perturbation $\Bar{\mathbf{E}}^\mathbf{W}$;
    %\STATE Compute the perturbation $\Bar{\mathbf{E}}^\mathbf{B}=\frac{1}{s}\Bar{\mathbf{E}}^\mathbf{W}\mathbf{A}^+_1$ on matrix $\mathbf{B}$ according to Equation (14);
    \STATE EMA perturbation $\hat{\mathbf{E}}^\mathbf{B}_1 =  \Bar{\mathbf{E}}^{{\mathbf{B}}}_1$;
    \STATE Update the weights using the gradient obtained in Step 4;
    \STATE Update the parameters to the next perturbation point $(\mathbf{W}_0$,  $\mathbf{B}_1+\hat{\mathbf{E}}^{\mathbf{B}}_1$, $\mathbf{A}_1)$.
    \ELSE
    \STATE Calculate the gradient at the perturbation point $(\mathbf{W}_0$,  $\mathbf{B}_{t-1}+\hat{\mathbf{E}}^{\mathbf{B}}_{t-1}$, $\mathbf{A}_{t-1})$.
    %\STATE Apply Equation (12) to compute the gradient in the full parameter space ${\Bar{\mathbf{g}}^\mathbf{W}}$;
    %\STATE Use Equation (13) to calculate the perturbation $\Bar{\mathbf{E}}^\mathbf{W}$;
    \STATE Compute the perturbation $\Bar{\mathbf{E}}^\mathbf{B}_t=\frac{1}{s}\Bar{\mathbf{E}}^\mathbf{W}\mathbf{A}^+_{t-1}$ on matrix $\mathbf{B}$ according to Equation (14);
    \STATE Return to the original parameter point $(\mathbf{W}_0$,  $\mathbf{B}_{t-1}$, $\mathbf{A}_{t-1})$. 
    \STATE Calculate the EMA perturbation ${{\hat{\mathbf{E}} }^\mathbf{B}_t} = (1 - \beta ){{\hat{\mathbf{E}}}^\mathbf{B}_{t - 1}} + \beta {\Bar{\mathbf{E}}^\mathbf{B}_t}$.
    \STATE Update the weights to $(\mathbf{W}_0$,  $\mathbf{B}_t$, $\mathbf{A}_t)$ using the gradient obtained in Step 9;
    \STATE Update the parameters to the next perturbation point $(\mathbf{W}_0$,  $\mathbf{B}_t+\hat{\mathbf{E}}^{\mathbf{B}}_t$, $\mathbf{A}_t)$.
    \ENDIF
    \ENDFOR
    \end{algorithmic}
    \label{algorithm:2}
   % \end{small}
\end{algorithm}

\section{D. More experiments}

\subsection{D.1 The approximate ability of EMA perturbation}
We consider few shot learning with LoRA on RoBERTa-large. 
Fig.~\ref{fig_sema} illustrates the evolution of the difference in sharpness, $\left[ {L({\mathbf{w}_t} + {\bm{\hat{\varepsilon} }_{t}}) - L({\mathbf{w}_t})} \right] - \left[ {L({\mathbf{w}_t} + {\bm{\tilde \varepsilon }_t}) - L({\mathbf{w}_t})} \right]$, as described in Theorem 2, during training on six datasets (SNLI, SST-2, SST-5, MNLI, RTE, and TREC). $S^{\text{EMA}}$ denotes the sharpness computed using EMA perturbations, while $S^{\text{SAM}}$ refers to the original SAM sharpness. As training progresses, the absolute difference consistently decreases across all datasets, demonstrating that the EMA perturbation becomes increasingly effective at approximating the SAM perturbations. This validates the use of EMA perturbations as a computationally efficient surrogate for SAM perturbations. This result empirically supports Theorem 2.

\subsection{D.2 The change in balancedness during FMLoRA training}
We consider few shot learning with LoRA on RoBERTa-large. For dataset MNLI, 1st, 12th and 24th query layers’ $2|B_{t,l}|$ are
plotted, where $t$ denotes the iteration and $l$ denotes the layer index. The layers are chosen to represent early, middle, and final stages of RoBERTa. Balancedness of FMLoRA and Adam on different layers are plotted in Fig.~\ref{fig_banan}. Balancedness may increase or decrease across different layers. As shown in Fig.~\ref{fig_banan}, the balancedness of FMLoRA in the first query layer of RoBERTa-large gradually decreases during training, while in the 12th layer, it first decreases and then increases. In contrast, the balancedness in the 24th layer continuously increases. An increase typically occurs when parameter magnitudes in both low-rank subspaces grow simultaneously. This behavior can be influenced by factors such as the learning rate, optimization algorithm, weight decay, and other regularization strategies. Despite these occasional increases, FMLoRA generally maintains lower balancedness than Adam in most layers, suggesting its capacity to induce implicit regularization during training.

\bibliography{aaai2026}

\end{document}